\documentclass[journal]{IEEEtran}
\ifCLASSINFOpdf
\else
\fi

\usepackage{times}
\usepackage{epsfig}
\usepackage{graphicx}
\usepackage{amsmath}
\usepackage{amssymb}
\usepackage{multirow}
\usepackage{multicol}
\usepackage{float}
\usepackage{caption}
\usepackage{subcaption}
\newcommand{\tabincell}[2]{\begin{tabular}{@{}#1@{}}#2\end{tabular}}
\usepackage[breaklinks=true,bookmarks=false]{hyperref}

\begin{document}
%
\title{Spectral Regularization for Combating Mode Collapse in GANs}
%
%
%

\author{Kanglin~Liu,
        Wenming~Tang,
        Ruitao~Xie,
        and~Guoping~Qiu \\
        \captionsetup{type=figure}\setcounter{figure}{0}
        \includegraphics[width=.24\textwidth]{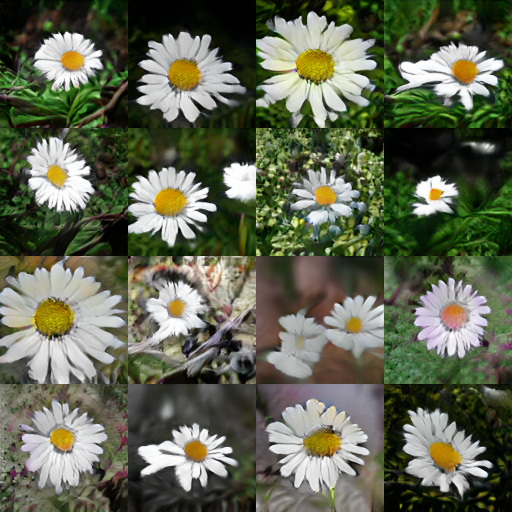}
        \includegraphics[width=.24\textwidth]{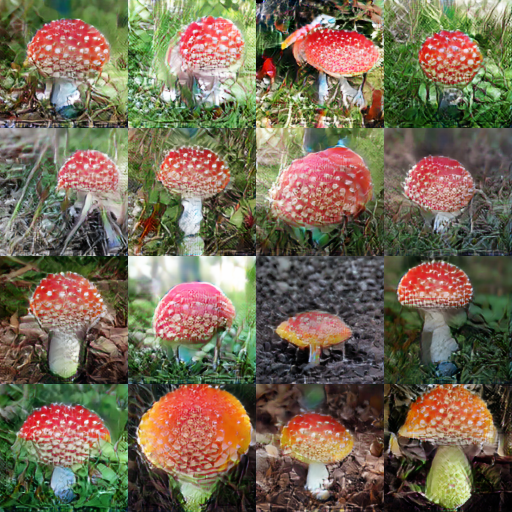}
        \includegraphics[width=.24\textwidth]{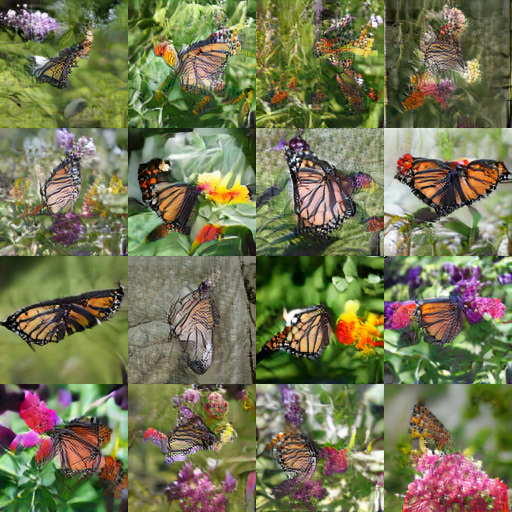}
        \includegraphics[width=.24\textwidth]{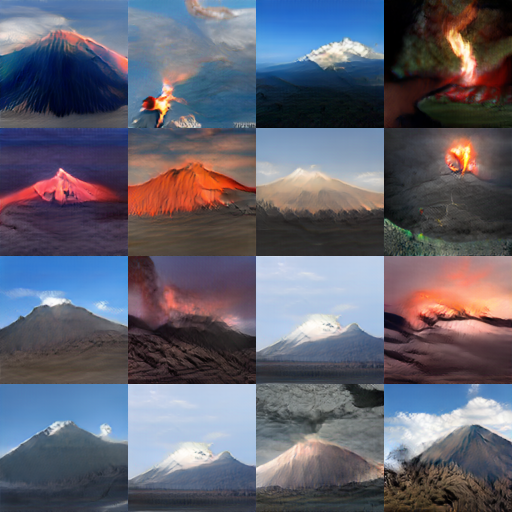}
        \label{fig:fig}
        \captionof{figure}{Conditional samples generated by our model base on ImageNet with setting $E_{2048-64}$ in Table \ref{t11}. With the same setting, SN-GANs suffer from mode collapse, which starts at iteration = 40k. On the contrary, our model can gain performance improvement steadily with iterations, and no mode collapse occurs.}
        \vspace*{-1cm}
\thanks{K. Liu is with the Department
of Information Engineering, Shenzhe University, Shenzhen,
518061, China. e-mail: max.liu.426@gmail.com.}
\thanks{G. Qiu is with University of Nottingham and Shenzhen University.}
\thanks{Manuscript received **, 2019; revised**, **.}}

%
%

\markboth{Published as a conference paper on ICCV 2019}%
{Shell \MakeLowercase{\textit{et al.}}: Bare Demo of IEEEtran.cls for IEEE Journals}
%



%
\maketitle



\begin{abstract}
Despite excellent progress in recent years, mode collapse remains a major unsolved problem in generative adversarial networks (GANs).
In this paper, we present spectral regularization for GANs (SR-GANs), a new and robust method for combating the mode collapse problem in GANs. 
Theoretical analysis shows that the optimal solution to the discriminator has a strong relationship to the spectral distributions of the weight matrix.
Therefore, we monitor the spectral distribution in the discriminator of spectral normalized GANs (SN-GANs), and discover a phenomenon which we refer to as spectral collapse, where a large number of singular values of the weight matrices drop dramatically when mode collapse occurs. 
We show that there are strong evidences linking mode collapse to spectral collapse; and based on this link, we set out to tackle spectral collapse as a surrogate of mode collapse. We have developed a spectral regularization method where we compensate the spectral distributions of the weight matrices to prevent them from collapsing, which in turn successfully prevents mode collapse in GANs. We provide theoretical explanations for why SR-GANs are more stable and can provide better performances than SN-GANs. We also present extensive experimental results and analysis to show that SR-GANs not only always outperform SN-GANs but also always succeed in combating mode collapse where SN-GANs fail.
The code is available at \url{https://github.com/max-liu-112/SRGANs}
\end{abstract}

\begin{IEEEkeywords}
Generative Adversarial Networks, Spectral Regularization, Mode Collapse, Lipschitz Constraint.
\end{IEEEkeywords}

%
\IEEEpeerreviewmaketitle

\section{Introduction}

Generative Adversarial Networks (GANs) \cite{A1} are one of the most significant developments in machine learning research of the past decade. Since their first introduction, GANs have attracted intensive interest in the machine learning community not only for their ability to learn highly structured probability distributions but also for their theoretically implications  \cite{A1,A2,A3,A4}. Essentially, GANs are constructed around two functions \cite{A5, A6}: the generator $\mathbf{G}$, which maps a sample \textit{z} to the data distribution, and the discriminator $\mathbf{D}$, which is trained to distinguish real samples of a dataset from fake samples produced by the generator. With the goal of reducing the difference between the distributions of generated and real samples, a GAN training algorithm trains $\mathbf{G}$ and $\mathbf{D}$ in tandem. 

GAN training is dynamic and sensitive to nearly every aspect of its setup, from optimization parameters to model architecture \cite{A7}. Training instability, or mode collapse, is one of the major obstacles in developing applications. Despite excellent progresses in recent years \cite{A8,A9,A10,A11,A12}, the mode collapse problem still persists. For example, one of the most impressive works to emerge recently is BigGANs \cite{A7}, which is the largest published GAN system based on the state of the art Spectral Normalization (SN-GAN)\cite{A10}. However, BigGANs can still suffer from the training instability problem, especially when the batch size is scaled up. Although implementing training stabilization measures such as employing $R_{1}$ zero-centred gradient penalty term \cite{A7} in the loss metric of the discriminator to prevent spectral noise can improve stability, this can cause severe degradation in performance, resulting in a 45\% reduction in Inception Score. 

In this paper, we present Spectral Regularization, a robust method for combating the mode collapse problem in GANs. Theoretically, we have found that spectral distribution is closely related to $D$'s performance, and affect how the supremum  of Lipschitz constraint can be reached. Regarding mode collapse as the rapid decay in performance, we reason that the spectral distributions in $\mathbf{D}$ may also have a strong relation to mode collapse. 
Through comprehensive analysis of spectral distributions in a large number of GAN models trained with the state of the art SN-GAN algorithm, we discover that when mode collapse occurs to a model, spectral distributions of $\overline{W}_{\mathrm{SN}}(W)$ in $\mathbf{D}$ also collapse, where $\overline{W}_{\mathrm{SN}}(W)$ is spectral normalized weight matrix. Specifically, we observe that when a model performs well and no mode collapse occurs, there are a large number of singular values of $\overline{W}_{\mathrm{SN}}(W)$ in $\mathbf{D}$ very close to 1, and that when mode collapse occurs to a model, singular values of $\overline{W}_{\mathrm{SN}}(W)$ in  $\mathbf{D}$ will drop dramatically. We refer to the phenomenon where a large number of singular values drop significantly as {\bf spectral collapse}. 

In all GAN models of various sizes and trained with a variety of parameter settings on datasets extensively used in the literature, we observe that mode collapse and spectral collapse always go side by side. This fact leads us to reason that mode collapse in SN-GANs is caused by spectral collapse in  $\mathbf{D's}$ weight matrices. Based on such insight into spectral distributions of $\overline{W}_{\mathrm{SN}}(W)$, we propose a new and robust method called spectral regularization to prevent GANs from mode collapse.  In addition to normalizing the weight matrices, spectral regularization imposes constraints on $\mathbf{D's}$ weight matrices by compensating their spectral distributions to avoid spectral collapse. Theoretical analysis shows that spectral regularization is better than spectral normalization at preventing weight matrix from concentrating into one particular direction.
We show that SN-GANs are a special case of spectral regularization, and in a series of extensive experiments we demonstrate that spectral regularization  not only provides superior performances to spectral normalization but also can always avoid mode collapse in cases where spectral normalization failed. 

Our contributions can be summarized as follows:

(1) Through theoretical analysis and extensive experimental observations, we provide an insight into the likely causes of mode collapse in a state of the art GAN normalization technique, spectral normalization (SN-GANs). We introduce the concept of {\bf spectral collapse} and provide strong evidence to link spectral collapse with mode collapse in SN-GANs. 

(2) Based on above insight, we have developed a new robust regularization method, {\bf Spectral Regularization}, where we compensate the spectral distributions of the weight matrices in $\mathbf{D}$ to prevent spectral collapse, thus preventing mode collapse in GANs. Extensive experimental results show that spectral regularization not only can always prevent mode collapse 
but also can consistently provide improved performances over SN-GANs.

%
%
%
%

\section{Analysis of Mode Collapse in SN-GANs}\label{sec2}
\subsection{A Brief Summary of SN-GANs}
For easy discussion, we first briefly recap the essential ideas of the spectral normalization technique for training GANs \cite{A10}. As far we are aware, this is currently one of the best methods in the literature and has been successfully used to construct large systems such as BigGANs \cite{A7} .  For convenience, we largely follow the notation convention of \cite{A10}. Considering a simple discriminator of a neural network of the following form:
\begin{equation}\label{key1}
\setlength{\abovedisplayskip}{3pt}
\setlength{\belowdisplayskip}{3pt}
f(x,\theta)=W^{L+1}(a_{L} \cdot W^{L} \cdot a_{L-1} \cdot W^{L-1} \cdots a_{1}W^{1}x)
\end{equation}
where $ \theta :=\{W^{1}, \cdots ,W^{L}, W^{L+1}\}$ is the learning parameters set, $ W^{l} \in \mathbb{R}^{d_{l} \times d_{l-1}} $, $ W^{L+1} \in \mathbb{R}^{1 \times d_{L}} $, and $ a_{l} $ is an element-wise non-linear activation function. We omit the bias term of each layer for simplicity. The final output of the discriminator is given by
\begin{equation}\label{key2}
\setlength{\abovedisplayskip}{3pt}
\setlength{\belowdisplayskip}{3pt}
D(x, \theta)=\mathcal{A}(f(x, \theta))
\end{equation}
where $\mathcal{A}$ is an activation function corresponding to the divergence of a distance measure of users' choice.

The standard formulation of GANs is given by \cite{A10,A2}:
\begin{equation}\label{key3}
\setlength{\abovedisplayskip}{3pt}
\setlength{\belowdisplayskip}{2pt}
\underset{G}{\mathrm{min}} \ \underset{D}{\mathrm{max}} V(G, D)
\end{equation}	
where min and max of \textit{G} and \textit{D} are taken over the set of the generator and discriminator functions respectively. The conventional form of \textit{V}(\textit{G}, \textit{D}) is given by $ E_{x \sim q_{data}}[\mathrm{log}D(x)]+E_{x'\sim q_{G}}[\mathrm{log}(1-D(x'))]$ \cite{A10}, 
where $ q_{data}$ is the data distribution and $ q_{G}$ is the model (generator) distribution. 

To guarantee Lipschitz continuity, spectral normalization \cite{A10} controls the Lipschitz constant of the discriminator function by literally  constraining the spectral norm of each layer:
\begin{equation}\label{key4}
\setlength{\abovedisplayskip}{3pt}
\setlength{\belowdisplayskip}{3pt}
\overline{W}_{\mathrm{SN}}(W):=W/\sigma(W)
\end{equation}
where $\sigma$(W) is the spectral norm of the weight matrix \textit{W} in the discriminator network, which is equivalent to the largest singular value of \textit{W}.

The authors of SN-GANs \cite{A10} and those of BigGANs \cite{A7} have demonstrated the superiority of spectral normalization over other normalization or regularization techniques, e.g., gradient penalty \cite{A8}, weight normalization\cite{A11} and orthonormal regularization \cite{A19}. However, as a state of the art GAN model, BigGANs (based on spectral normalization) can still suffer from mode collapse.
Therefore, mode collapse remains an unsolved open problem, seeking better and more robust solution is very important for advancing GANs.

\subsection{Theoretical Analysis}

Our theoretical analysis has found that the spectral distribution of the weight matrix determine how the supremum of Lipschitz constraint can be reached, and affect performance of $D$.

To be specific. SN-GANs enforce Lipschitz constraint on the weight matrix in each convolutional layers of $D$ by constraining the spectral norm, thus guaranteeing $D$ satisfy Lipschitz constraint. Furthermore, we have found that the spectral distribution affects how the supremum of Lipschitz constraint can be reached. To verify this, Corollary 1 is present here. (see proof in Appendix)

\noindent \textbf{Corollary 1.} If a linear function $f = Wx$ satisfies Lipschitz constraint: $\left \| f(x_1) - f(x_2) \right \| \leq \left \| x_1 - x_2 \right \| $, then the supremum of the Lipschitz constraint is obtained when all the singular values of the weight matrix $W$ are 1.

On the other hand, through singular value decomposition, we find that spectral distribution affect the performance of $D$.

The weight matrix $W$ in the convolutional layers can be expressed as:
\begin{equation}\label{key66}
\setlength{\abovedisplayskip}{3pt}
\setlength{\belowdisplayskip}{3pt}
\begin{split}
W&=U\cdot \Sigma \cdot V^{T}\\
&=\sigma_{1}u_1v_1+\sigma_{2}u_2v_2+\cdots+\sigma_{n}u_nv_n
\end{split}
\end{equation}
where both \textit{U} and \textit{V} are orthogonal matrix, the columns of \textit{U},  $\left [ u_{1}, u_{2}, \cdots, u_{m} \right ]$, are called left singular vectors of \textit{W}, the columns of \textit{V},  $\left [ v_{1}, v_{2}, \cdots, v_{n} \right ]$,  are called right singular vectors of \textit{W}, and $\sigma \cdots \sigma_{n}$ are the singular values of $W$.

\begin{table}[htp]
	\caption{Experiment settings. The experiments are divided into 5 groups $A, B, C, D$ and $E$. Within each group, the models share exactly the same network architecture but differ in batch size. For groups $A-D$, we vary the batch sizes inside each group to study how batch sizes relate to mode collapse, and we change the channel sizes between groups to investigate how discriminator capacity affects mode collapse.  Group $E$ is experiments applied to a different data set. The purpose is to evaluate how different data affect mode collapse. Batch represents the batch size. CH is the channel size of the discriminator. The subscript of each group name annotates the batch and channel setting of that experiment, e.g., $A_{a-b}$ represents setting with a batch size $a$ and a CH size $b$.}
	\centering
	\setlength{\abovecaptionskip}{-0.2cm} 
	\setlength{\belowcaptionskip}{0cm} 
	\vspace*{-0.1cm}
	\resizebox{90mm}{25mm}{
		\begin{tabular}{l c c c |l c c c }
			\hline \hline
			Setting & Batch & CH & Dataset&Setting & Batch & CH & Dataset \\ 	
			\hline
			$A_{16-128}$ & 16 & 128  &  CIFAR-10&$C_{8-32}$ & 8 & 32  & CIFAR-10\\ 
			$A_{32-128}$ & 32 & 128  &  CIFAR-10&$C_{16-32}$ & 16 & 32  &  CIFAR-10 \\ 
			$A_{64-128}$ & 64 & 128  &  CIFAR-10&$C_{32-32}$ & 32 & 32  &  CIFAR-10 \\ 
			$A_{128-128}$ & 128 & 128  &  CIFAR-10&$C_{64-32}$ & 64 & 32  &  CIFAR-10  \\ 
			\cline{5-8}
			$A_{256-128}$ & 256 & 128 &  CIFAR-10&$D_{128-256}$& 128 & 256  &  CIFAR-10  \\ 
			$A_{512-128}$ & 512 & 128 &  CIFAR-10&$D_{256-256}$ & 256 & 256  & CIFAR-10 \\ 
			$A_{1024-128}$ & 1024 & 128 &  CIFAR-10&$D_{512-256}$ & 512 & 256  &  CIFAR-10 \\ 
			\cline{1-8}
			$B_{8-64}$ & 8 & 64 &  CIFAR-10& $E_{64-128}$& 64 & 128  &   STL-10 \\ 
			$B_{16-64}$ & 16 & 64  &  CIFAR-10 &$E_{256-128}$ & 256 & 128  & STL-10  \\ 
			$B_{32-64}$ & 32 & 64  &  CIFAR-10 &$E_{256-64}$ & 256 & 64 & STL-10 \\ 
			$B_{64-64}$ & 64 & 64  &  CIFAR-10& $E_{256-32}$& 256 & 32  & STL-10\\ 
			$B_{128-64}$ & 128 & 64  &  CIFAR-10 & $E_{512-64}$ & 512 & 64  &  ImageNet\\ 
			$B_{256-64}$ & 256 & 64  &  CIFAR-10 & $E_{2018-64}$ & 2048 & 64  &  ImageNet \\ 
			\hline \hline
	\end{tabular} }
	\vspace*{-0.0cm}
	\label{t11}
\end{table}

We can see that, $\sigma$ in Equation \ref{key66} determine how corresponding singular vectors are utilized in implementing convolutional operation $Wx$. Extremely, if $\sigma_{1} > 0$, and $\sigma_{2}=\cdots=\sigma_{n}=0$, we can see that the convolutional operation $Wx$ only uses the first singular value and its corresponding singular vector.

Through the theoretical analysis above, it is clear that spectral distribution plays a key role in determining Lipschitz constraint and $D$'s performance. However, spectral normalization only constrains the spectral norm, and takes no consideration on spectral distribution. 
Recalling that mode collapse can be regarded as the intense decay in $D$'s performance, we reason that mode collapse may also have a strong relationship with spectral distribution. Furthermore, finding such a relationship may helps solve the mode collapse problem.

\subsection{Mode Collapse vs Spectral Collapse}
In order to find the likely link between mode collapse and spectral distributions, we have conducted a series of experiments for unconditional image generation on  CIFAR-10 \cite{A13}, STL-10 \cite{A17} and conditional image generation on ILRSVRC2012 \cite{A21} datasets.  Our implementation is based on the SN-GANs architecture of  \cite{A10}, which uses the hinge loss as the discriminator objective and is given by:
\begin{equation}\label{key5}
\setlength{\abovedisplayskip}{3pt}
\setlength{\belowdisplayskip}{3pt}
\begin{split}
L_{D}=&E_{x \sim q_{data}}[\mathrm{min}(0,-1+D(x))]\\
&+E_{x \sim q_{G}}[\mathrm{min}(0,-1-D(x))]
\end{split}
\end{equation}

The optimization settings follow  literature \cite{A10, A18}.
Previous authors  have shown that increasing  batch size or decreasing  discriminator capacity could potentially lead to mode collapse \cite{A7}. We therefore conduct experiments for various combinations of batch and channel sizes as listed in Table \ref{t11}. We follow the practices in the literature of using Inception Score (IS) \cite{A16} and Fr\'{e}chet Inception Distance (FID) \cite{A17} as approximate measures of sample quality, and results are shown in Table \ref{t22} where we also identify all settings where mode collapse has occurred to SN-GANs. Through monitoring Inception Scores, Fr\'{e}chet Inception Distance and synthetic images during training, mode collapse is observed in 10 settings including $B_{64-64}$,  $B_{128-64}$, $B_{256-64}$, $C_{8-32}$, $C_{16-32}$, $C_{32-32}$, $C_{64-32}$, $E_{256-64}$, $E_{256-32}$ and $E_{2048-64}$. In other 16 settings, mode collapse has not happened.  

Mode collapse is a persistent problem in GAN training and is also a major issue in SN-GANs as has been shown in BigGANs\cite{A7} and in Table \ref{t22}.  Here, we monitor the entire spectral distributions of SN-GANs, i.e., all singular values of $\overline{W}_{\mathrm{SN}}(W)$ in the discriminator network during training.

The discriminator network in our implementation uses the same architecture as that in the original SN-GANs\cite{A10}. For image generation on CIFAR-10 and STL-10, there are 10 convolutional layers. As for image generation on ImageNet, 17 convolutional layers are included. Please see Appendix for the setting details. In order to discover the likely link of mode collapse to spectral distribution, we plot the spectral distributions of every layer of the discriminator for all 26 settings. In the following, we present some typical examples and readers are referred to the Appendix for  all other plots.

\begin{figure*}
	\begin{subfigure}{.19\textwidth}
		\centering
		\includegraphics[width=1\linewidth]{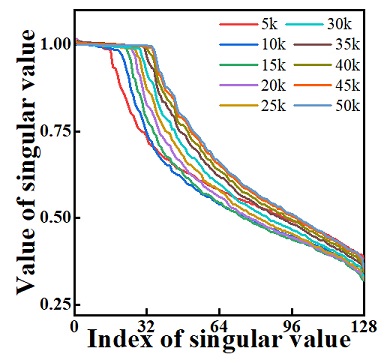}  
		\vspace*{-0.6cm}
		\caption{\textrm{\small $A_{256-128}$}}
	\end{subfigure}
	\begin{subfigure}{.19\textwidth}
		\centering
	\includegraphics[width=1\linewidth]{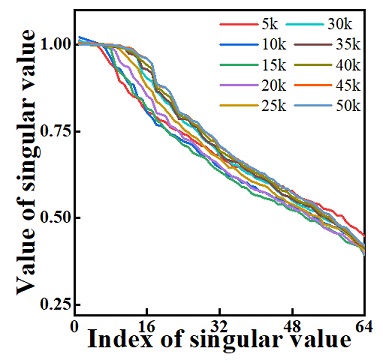}  
		\vspace*{-0.6cm}
		\caption{\textrm{\small $B_{32-64} $}}
	\end{subfigure}
	\begin{subfigure}{.19\textwidth}
		\centering
		\includegraphics[width=1\linewidth]{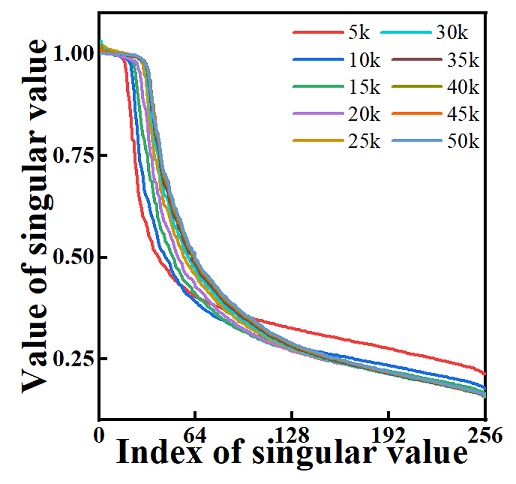}  
		\vspace*{-0.6cm}
		\caption{\textrm{\small $D_{128-256} $}}
	\end{subfigure}
	\begin{subfigure}{.19\textwidth}
		\centering
		\includegraphics[width=1\linewidth]{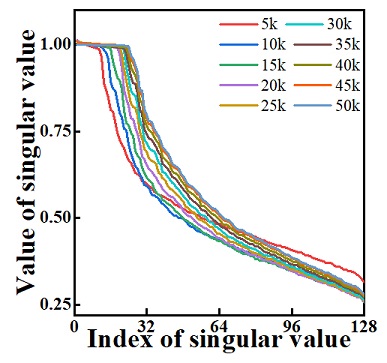}
		\vspace*{-0.6cm}
		\caption{\textrm{\small  $E_{256-128} $}}	
	\end{subfigure}
	\begin{subfigure}{.19\textwidth}
		\centering
		\includegraphics[width=1\linewidth]{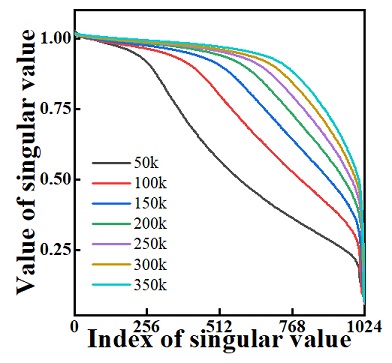}
		\vspace*{-0.6cm}	
		\caption{\textrm{\small $E_{512-64} $}	}	
	\end{subfigure}
	\vspace*{-0.0cm}
	\caption{Spectral distributions in the last layer for Good GANs (no mode collapse) at different number of iterations. The curves represent the spectral distributions after $5k$ iterations, $10k$ iterations, ..., $50k$ iterations, ... . }
	\label{fig:fig1}
	\vspace*{-0.0cm}
\end{figure*}

\begin{figure*}
	\begin{subfigure}{.19\textwidth}
		\centering
		\includegraphics[width=1\linewidth]{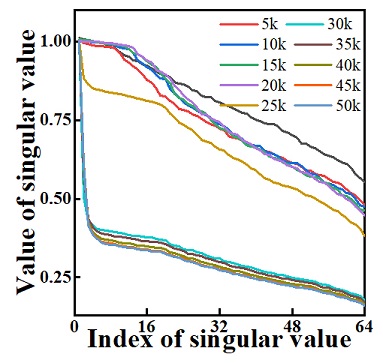}  
		\vspace*{-0.6cm}
		\caption{\textrm{\small $B_{64-64}$}}
	\end{subfigure}
	\begin{subfigure}{.19\textwidth}
		\centering
		\includegraphics[width=1\linewidth]{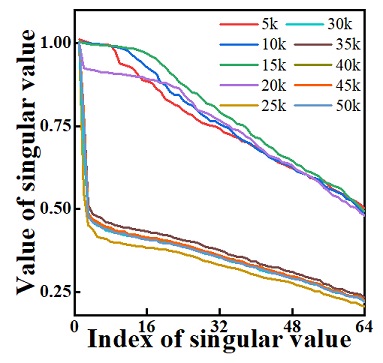}  
		\vspace*{-0.6cm}
		\caption{\textrm{\small $B_{128-64}$}}
	\end{subfigure}
	\begin{subfigure}{.19\textwidth}
		\centering
		\includegraphics[width=1\linewidth]{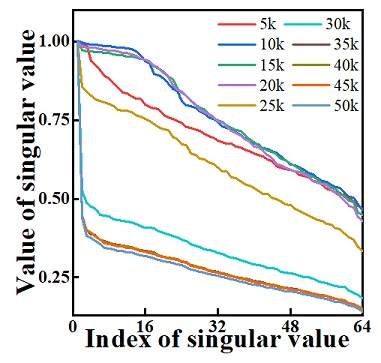}  
		\vspace*{-0.6cm}
		\caption{\textrm{\small $B_{256-64}$}}
	\end{subfigure}
	\begin{subfigure}{.19\textwidth}
		\centering
		\includegraphics[width=1\linewidth]{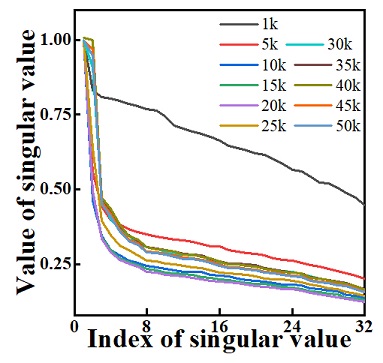}
		\vspace*{-0.6cm}
		\caption{\textrm{\small  $C_{8-32} $}}	
	\end{subfigure}
	\begin{subfigure}{.19\textwidth}
		\centering
		\includegraphics[width=1\linewidth]{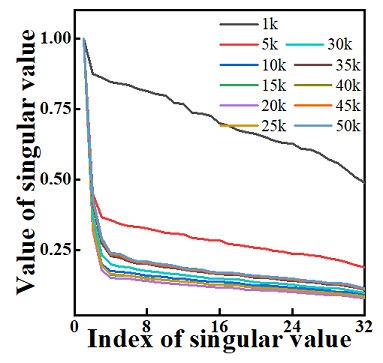}
		\vspace*{-0.6cm}	
		\caption{\textrm{\small $C_{16-32} $}	}	
	\end{subfigure}
	\newline
	\begin{subfigure}{.19\textwidth}
		\centering
		\includegraphics[width=1\linewidth]{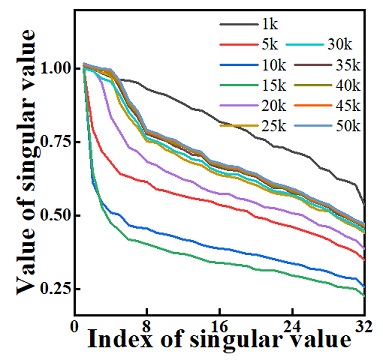}  
		\vspace*{-0.6cm}
		\caption{\textrm{\small $C_{32-32}$}}
	\end{subfigure}
	\begin{subfigure}{.19\textwidth}
		\centering
		\includegraphics[width=1\linewidth]{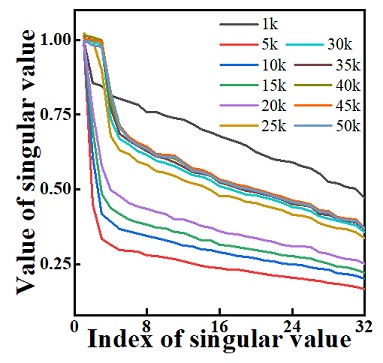}  
		\vspace*{-0.6cm}
		\caption{\textrm{\small $C_{64-32}$}}
	\end{subfigure}
	\begin{subfigure}{.19\textwidth}
		\centering
		\includegraphics[width=1\linewidth]{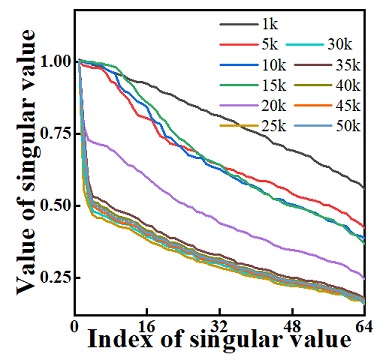}  
		\vspace*{-0.6cm}
		\caption{\textrm{\small $E_{256-64} $}}
	\end{subfigure}
	\begin{subfigure}{.19\textwidth}
		\centering
		\includegraphics[width=1\linewidth]{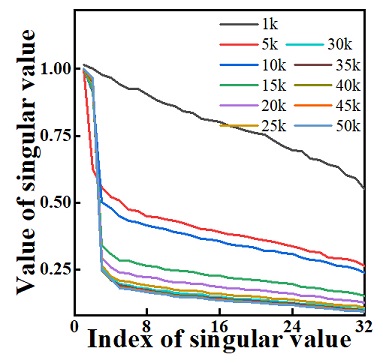}
		\vspace*{-0.6cm}
		\caption{\textrm{\small  $E_{256-32} $}}	
	\end{subfigure}
	\begin{subfigure}{.19\textwidth}
		\centering
		\includegraphics[width=1\linewidth]{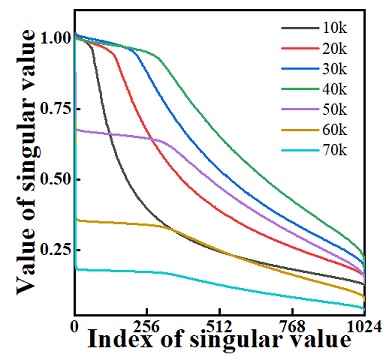}
		\vspace*{-0.6cm}	
		\caption{\textrm{\small $E_{2048-64} $}}	
	\end{subfigure}
	\vspace*{-0.0cm}
	\caption{Spectral distributions in the last layer for settings where mode collapse occurs. The curves represent the spectral distributions after $1k$ iterations, $10k$ iterations, ..., $50k$ iterations, ... .}
	\label{fig:fig2}
	\vspace*{0.4cm}
	\begin{subfigure}{.19\textwidth}
		\centering
		\includegraphics[width=1\linewidth]{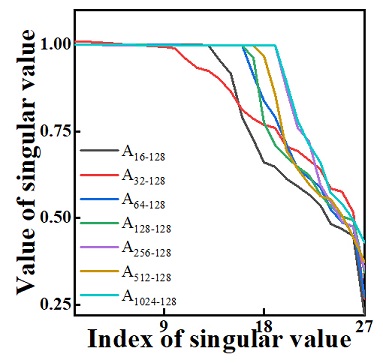}  
		\vspace*{-0.4cm}
		\caption{\textrm{\small  group $A$}}
	\end{subfigure}
	\begin{subfigure}{.19\textwidth}
		\centering
		\includegraphics[width=1\linewidth]{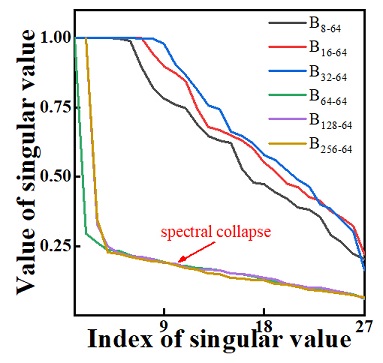}  
		\vspace*{-0.4cm}
		\caption{\textrm{\small  group $B$}}
	\end{subfigure}
	\begin{subfigure}{.19\textwidth}
		\centering
		\includegraphics[width=1\linewidth]{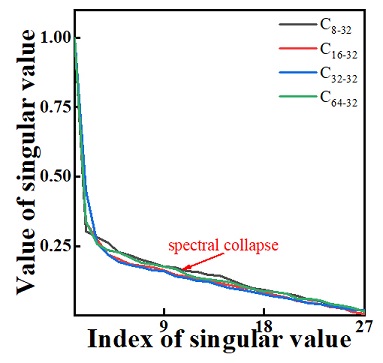}  
		\vspace*{-0.4cm}
		\caption{\textrm{\small  group $C$}}
	\end{subfigure}
	\begin{subfigure}{.19\textwidth}
		\centering
		\includegraphics[width=1\linewidth]{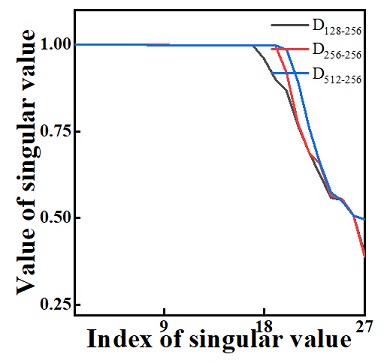}
		\vspace*{-0.4cm}
		\caption{\textrm{\small  group $D$}}	
	\end{subfigure}
	\begin{subfigure}{.19\textwidth}
		\centering
		\includegraphics[width=1\linewidth]{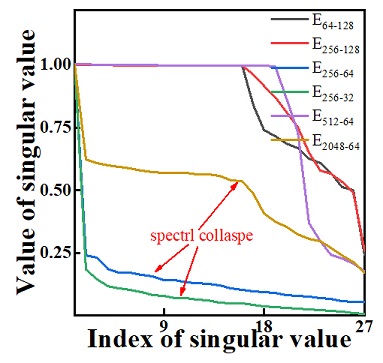}
		\vspace*{-0.4cm}	
		\caption{\textrm{\small  group $E$}}	
	\end{subfigure}
	\caption{Spectral distributions (after $50k$ iterations) in $layer\_0$ for different settings.}
	\label{fig:fig3}
	
\end{figure*}

\begin{figure*}
	\begin{subfigure}{.32\textwidth}
		\centering
		\includegraphics[width=0.8\linewidth]{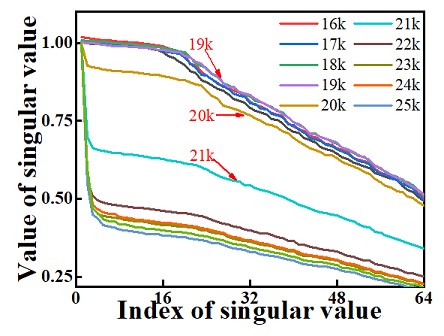}  
		\vspace*{-0.2cm}
		\caption{\textrm{\small spectral distributions }}
	\end{subfigure}
	\begin{subfigure}{.32\textwidth}
		\centering
		\includegraphics[width=0.8\linewidth]{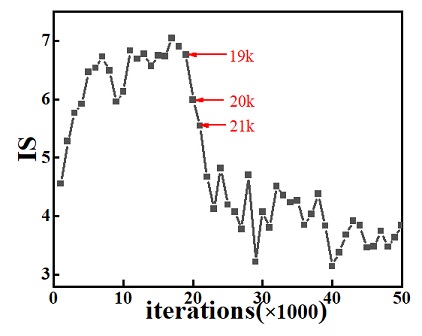}  
		\vspace*{-0.2cm}
		\caption{\textrm{\small  Inception Score }}
	\end{subfigure}
	\begin{subfigure}{.32\textwidth}
		\centering
		\includegraphics[width=0.8\linewidth]{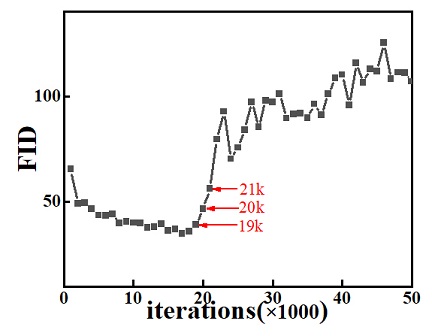}  
		\vspace*{-0.2cm}
		\caption{\textrm{\small Fr\'{e}chet Inception Distance}}
	\end{subfigure}
	\caption{An example showing how spectral distributions relate to Inception Score and Fr\'{e}chet Inception Distance. Here the setting is  $B_{128-64}$ and the spectral distributions correspond to those of $layer\_9$.}
	\label{fig:fig31}
\end{figure*}

Figure \ref{fig:fig1} shows the spectral distributions of the last layer ($layer\_9$ for task on CIFAR-10 and STL-10, $layer\_16$ for task on ImageNet) of 5 settings where mode collapse does not happen. Figure \ref{fig:fig2} shows the spectral distributions of the last layer of all 10 settings where mode collapse has occurred.  Through analyzing the spectral distribution plots in Figure \ref{fig:fig1} and Figure \ref{fig:fig2}, we notice a very interesting pattern. In the cases where no mode collapse happens, the shapes of the spectral distribution curves do not change significantly with the number of training iteration. On the other hand, for those settings where mode collapse has occurred, the shapes of the spectral distribution curves change significantly as training progresses. In particular, a large number of singular values become very small when training passes a certain number of iterations. This is as if the curves have "collapsed", and we refer to this phenomenon as {\bf spectral collapse}. 

The phenomenon of spectral collapse is also observed across different settings. Figure \ref{fig:fig3} plots the spectral distributions of the 5 groups of experimental settings in Table \ref{t11}. It is seen that in groups $A$ and $D$, the spectral distributions across different settings are very similar and no spectral collapse is observed. Very interestingly, no mode collapse is observed either. In group $B$, the spectral distributions of $B_{64-64}, B_{128-64}$ and $B_{256-64}$ have collapsed, not surprisingly, mode collapse also happens to these 3 settings. In group $C$, the spectral distributions of all settings have collapsed, i.e., most singular values are very small (except for the first one which is forced to be 1 by spectral normalization). Again as expected, mode collapse happens to all settings in this group. In group $E$, it is seen that settings $E_{256-64}$, $E_{256-32}$ and $E_{2048-64}$ have suffered from spectral collapse. Again, mode collapse is observed for these three settings. 


In order to understand what has happened when spectral collapse occurs, Figure \ref{fig:fig31} shows how a typical spectral distribution relates to Inception Score and  Fr\'{e}chet Inception Distance during training. It is seen that up to 19k iterations both IS and FID are showing good performances, and the corresponding spectral distribution has a large number of large singular values. At 20k iterations, IS and FID performances start to drop, correspondingly, the spectral distribution starts to fall. At 21k iterations, the IS and FID performances have dropped significantly and mode collapse has started, and very importantly, the spectral distribution has dropped dramatically - starting to collapse. 

The association of mode collapse with spectral collapse is observed for all the layers and on all settings (readers are referred to the Appendix for more examples). We therefore believe that mode collapse and spectral collapse happen at the same time, and spectral collapse is the likely cause of mode collapse. In the following section, we will introduce spectral regularization to prevent spectral collapse thus avoiding mode collapse.

\section{Spectral Regularization}
We have now established that spectral collapse is closely linked to mode collapse in SN-GANs. In this section, we introduce spectral regularization, a technique for preventing spectral collapse. We show that preventing spectral collapse can indeed solve the mode collapse problem, thus demonstrating that spectral collapse is the cause of mode collapse rather than a mere symptom.

Performing singular value decomposition, the weight matrix $ W $ can be expressed as:
\begin{equation}\label{key6}
\setlength{\abovedisplayskip}{3pt}
\setlength{\belowdisplayskip}{3pt}
W=U\cdot \Sigma \cdot V^{T}
\end{equation}
where both \textit{U} and \textit{V} are orthogonal matrix, the columns of \textit{U},  $\left [ u_{1}, u_{2}, \cdots, u_{m} \right ]$, are called left singular vectors of \textit{W}, the columns of \textit{V},  $\left [ v_{1}, v_{2}, \cdots, v_{n} \right ]$,  are called right singular vectors of \textit{W}, and $\Sigma$ can be expressed as:
\begin{equation}\label{key7}
\setlength{\abovedisplayskip}{3pt}
\setlength{\belowdisplayskip}{3pt}
\Sigma =\begin{bmatrix}
D &0 \\ 
0 &0 
\end{bmatrix}
\end{equation} 
where $D=diag\left \{ \sigma_{1}, \sigma_{2}, \cdots, \sigma_{r}  \right \}$ represents the spectral distribution of $W$.

When mode collapse occurs, spectral distributions concentrate on the first singular value, and the rest singular values drop dramatically (spectral collapse). To avoid spectral collapse, we can apply $\Delta D$ to compensate $D$. Here, we introduce two methods for implementation of $\Delta D$.
The first method is referred to as static compensation,  where $\Delta D$ is given by $diag\left \{ \sigma_{1}-\sigma_{1}, \sigma_{1}-\sigma_{2}, \cdots, \sigma_{1}-\sigma_{i},0,\cdots,0  \right \}$, and $i$ is a hyperparameter ($1 \le i\le r$).  In other words, static compensation is applied according to the difference between the first and $i$-th singular value. 

The other method is referred to as dynamic compensation. In the training process, we monitor the maximum ratio $\gamma_{j}= max(\frac{\sigma_{j}^{'}}{\sigma_{1}^{'}})$, where $\sigma_{j}^{'}$  represents recorded $j$-th singular value in the training process. Then, we utilize the difference between maximum ratio value and current value as the compensation. As a result, $\Delta D$ is given by $diag\left \{ 0, \gamma_{2}\cdot\sigma_{1}-\sigma_{2}, \cdots, \gamma_{r} \cdot\sigma_{1}-\sigma_{r} \right \}$.
Comparison of these two compensation method will be discussed in Section \ref{sec:exp}.

Spectral compensation turns $D$ into $D'$ as follows:  $D'=D+\Delta D$. Correspondingly, $W$ turns to $W'$: $W'=W+\Delta W$, where $\Delta W$ is given by:
\begin{equation}\label{key8}
\setlength{\abovedisplayskip}{3pt}
\setlength{\belowdisplayskip}{3pt}
\Delta W= U\cdot\begin{bmatrix}
\Delta D &0 \\ 
0 &0 \end{bmatrix} \cdot V^{T}=\sum_{k=2}^{N}(\Delta\sigma_{k})u_{k}v_{k}^{T}
\end{equation}
where $N$ is the number of singular value to be compensated, and $\Delta \sigma_{k}$ represents the compensation value in $k$-th singular value. For static compensation, $N$ in Equation \ref{key8} equals to the hyperparameter $i$, while for dynamic compensation, $N$ equals to the number of singular values $r$.

Finally, we apply spectral normalization to guarantee Lipschitz continuity, and obtain our spectral regularized $\overline{W}_{\mathrm{SR}}(W)$:
\begin{equation}\label{key9}
\setlength{\abovedisplayskip}{3pt}
\setlength{\belowdisplayskip}{3pt}
\overline{W}_{\mathrm{SR}}(W)=\frac{W+\Delta W}{\sigma(W)} =\overline{W}_{\mathrm{SN}}(W)+\Delta W/ \sigma(W)
\end{equation}
Clearly, spectral normalization is a special case of spectral regularization, when no compensation is applied.

\begin{table*}
	\centering
	\caption{IS and FID results for different settings, where IS is Inception Score and FID is Fr\'{e}chet Inception Distance. For IS, higher is better, while lower is better for FID. SN, SR represent Spectral normalization and Spectral Regularization, respectively.  MC stands for mode collapse, and SC stands for spectral collapse,  $\times$ represents that no mode collapse or spectral collapse occurs. \textbf{SN} in the MC column or SC column represents that mode collapse or  spectral collapse occurred to spectral normalization. Note that neither mode collapse nor spectral collapse happen to spectral regularization for all settings.}
	\resizebox{\textwidth}{28mm}{
		\begin{tabular}{l| c c|c c|c |c||l |c c| c c| c |c}
			\hline \hline
			\multirow{2}*{\tabincell{c}{Experiment\\ Setting}}&\multicolumn{2}{|c|}{IS}&\multicolumn{2}{c|}{FID}& \multirow{2}*{MC}&\multirow{2}*{SC}&\multirow{2}*{\tabincell{c}{Experiment\\ Setting}} &\multicolumn{2}{|c|}{IS}&\multicolumn{2}{c|}{FID}& \multirow{2}*{MC}&\multirow{2}*{SC}\\
			\cline{2-5}
			\cline{9-12}
			& SN& SR &SN& SR& & & &SN &SR& SN& SR & &   \\ 
			\hline
			$A_{16-128}$ & 8.15$\pm$.09 & 8.35$\pm$.09 &22.31$\pm$.28 &24.67$\pm$.28 & $\times$ & $\times$ &
			$C_{8-32}$   & 4.21$\pm$.18 & 4.93$\pm$.20 &80.00$\pm$1.12  &66.05$\pm$2.12 & \textbf{SN} & \textbf{SN} \\ 		
			
			$A_{32-128}$ & 8.38$\pm$.07 &8.45$\pm$.10 & 25.96$\pm$.42 & 22.00$\pm$.17 &$\times$ &$\times$   &   
			$C_{16-32}$  & 4.05$\pm$.15 &4.78$\pm$.23 & 79.69$\pm$.21&59.25$\pm$.43 & \textbf{SN} &  \textbf{SN} \\ 
			
			$A_{64-128}$ & 8.39$\pm$.15 & 8.65$\pm$.12 & 21.15$\pm$.15 &20.31$\pm$.18 & $\times$ & $\times$ &   
			$C_{32-32}$ & 4.29$\pm$.08  &4.70$\pm$.15  & 78.39$\pm$.17 &62.10$\pm$.24 & \textbf{SN}& \textbf{SN}  \\ 
			
			$A_{128-128}$ &8.61$\pm$.12 & 8.72$\pm$.08 & 21.01$\pm$.23  &19.98$\pm$.19  & $\times$&  $\times$  &  
			$C_{64-32}$ & 4.30$\pm$.14  &5.00$\pm$.14 & 85.15$\pm$1.20 &56.11$\pm$.54 & \textbf{SN}&   \textbf{SN}\\
			\cline{8-14}
			$A_{256-128}$ & 8.45$\pm$.14 &8.48$\pm$.03 & 20.87$\pm$.25  &19.87$\pm$.21   &$\times$ & $\times$  &  
			$D_{128-256}$ & 8.14$\pm$.06 & 8.92$\pm$.18 &24.43$\pm$.41 & 18.95$\pm$.23  & $\times$ &   $\times$\\ 	
		
			$A_{512-128}$ & 8.34$\pm$.09 &8.53$\pm$.04 & 21.85$\pm$.14 & 20.13$\pm$.12 & $\times$& $\times$   &  
			$D_{256-256}$ & 8.29$\pm$.12 & 8.83$\pm$.14& 22.54$\pm$.29 &19.56$\pm$.11  &$\times$ &$\times$  \\ 
			
			$A_{1024-128}$ & 8.31$\pm$.21 &8.52$\pm$.16  & 21.68$\pm$.35 &20.34$\pm$.13  &$\times$ &$\times$  & 
			 $D_{512-256}$&8.33$\pm$.09  &8.36$\pm$.12 & 22.58$\pm$.16 & 21.82$\pm$.29&$\times$ & $\times$ \\ 
			\cline{1-14}
			$B_{8-64}$ & 6.67$\pm$.05 &7.42$\pm$.06 & 45.19$\pm$.89 & 35.78$\pm$.11  &$\times$ &  $\times$& 
			 $E_{64-128}$& 8.98$\pm$.20 &9.14$\pm$.18  &42.40$\pm$.56  &39.89$\pm$.89  &$\times$ &$\times$ \\

			$B_{16-64}$ & 7.34$\pm$.06 &7.59$\pm$.08 & 31.73$\pm$.49& 29.42$\pm$.22 &$\times$ &  $\times$ & 
			$E_{256-128}$ & 9.10$\pm$.13 &9.11$\pm$.17 & 40.11$\pm$.89 &40.08$\pm$.29  & $\times$& $\times$ \\ 
			
			$B_{32-64}$ & 7.18$\pm$.03 &7.48$\pm$.09 & 33.76$\pm$.35 &28.60$\pm$.25  & $\times$&  $\times$ & 
			$E_{256-64}$ &7.38$\pm$.14  & 7.67$\pm$.06  &74.50$\pm$1.52  &69.20$\pm$.83    & \textbf{SN}&  \textbf{SN}\\ 
			
			$B_{64-64}$ & 6.96$\pm$.11 &7.52$\pm$.11 & 36.65$\pm$.29&  28.40$\pm$.36 & \textbf{SN}&   \textbf{SN} & 
			 $E_{256-32}$& 4.04$\pm$.11 & 4.38$\pm$.07  & 98.50$\pm$1.34 & 89.17$\pm$1.23 & \textbf{SN}& \textbf{SN}\\ 
			
			$B_{128-64}$ & 7.10$\pm$.14 & 7.13$\pm$.05 & 35.99$\pm$.48 &31.41$\pm$.56  & \textbf{SN}& \textbf{SN}  &
			 $E_{512-64}$& 34.13$\pm$.64 & \textbf{40.76$\pm$.73}& 59.88$\pm$.57 & \textbf{54.84$\pm$.68} &$\times$ &$\times$  \\ 
			
			$B_{256-64}$ & 6.85$\pm$.08 &7.58$\pm$.03 & 35.88$\pm$.42&27.68$\pm$.23 & \textbf{SN}&   \textbf{SN}&
			 $E_{2048-64}$& 21.78$\pm$.87 & \textbf{31.58 $\pm$.43}& 71.37$\pm$1.14 & \textbf{61.08 $\pm$.53} &\textbf{SN} &\textbf{SN}\\ 
			\hline \hline
	\end{tabular} }
	\vspace*{-0.0cm}
	\label{t22}
\end{table*}

\subsection{Gradient Analysis of Spectral Regularization}
We perform gradient analysis to show that spectral regularization provides a more effective way over spectral normalization in preventing $W$ from concentrating into one particular direction during training and thus avoiding spectral collapse. 

From Equation (\ref{key9}), taking static compensation as an example, we can write the gradient of $\overline{W}_{\mathrm{SR}}(W)$ with respect to $W_{ab}$ as:

\begin{equation}\label{key10}
\setlength{\abovedisplayskip}{1pt}
\setlength{\belowdisplayskip}{1pt}
\begin{split}
&\frac{\partial \overline{W}_{\mathrm{SR}}(W)}{\partial W_{ab}}=\frac{1}{\sigma(W)}\{E_{ab}-\overline{W}_{\mathrm{SN}}[u_{1}v_{1}^{T}]_{ab}\\
&-\frac{\Delta W}{\sigma(W)}[u_{1}v_{1}^{T}]_{ab}+\sum_{k=2}^{i}[u_{1}v_{1}^{T}-u_{k}v_{k}^{T}]_{ab}\cdot u_{k}v_{k}^{T}\}
\end{split}
\end{equation}
where $[\cdot]_{ab}$ represents the $(a,b)$-th entry of corresponding matrix, $E_{ab}$ is the matrix whose $(a,b)$-th entry is 1 and zero everywhere else.

We would like to comment on the implication of Equation (\ref{key10}). The first two terms, $E_{ab}-\overline{W}_{\mathrm{SN}}[u_{1}v_{1}^{T}]_{ab}$, are the gradient of spectral normalization $\frac{\partial \overline{W}_{\mathrm{SN}}(W)}{\partial{W_{ab}}}$ \cite{A10}, this is very easy to see from Equation (\ref{key9}). As explained in \cite{A10}, the second term can be regarded as being able to prevent the columns space of $W$ from concentrating into one particular direction in the course of training. In other words, spectral normalization prevents the transformation of each layer from becoming sensitive only in one direction. 
However, as we have seen  (e.g. Figure \ref{fig:fig2}), despite performing spectral normalization, the spectral distributions of $\overline{W}_{\mathrm{SN}}(W)$ can still concentrate on the first singular value thus causing spectral collapse. This shows the limited ability of spectral normalization in preventing $W$ from spectral collapse. 

In addition to the first two terms of spectral normalization, spectral regularization introduces the third and fourth terms in Equation (\ref{key10}). It can be seen that the third term enhances the effect of the second term, through which $W$ is much less likely to concentrate into one particular direction. Furthermore, the fourth term can be seen as the regularization term, encouraging $W$ to move along all $i$ directions pointed to by $u_{k}v_{k}^{T}$, for $k=1, 2, ..., i$, each weighted by the adaptive regularization coefficient $[u_{1}v_{1}^{T}-u_{k}v_{k}^{T}]_{ab}$. 
This encourages $W$ to make full use of the directions pointed to by $u_{j}v_{j}^{T}$, thus preventing $W$ from being concentrated on only 1 direction, which in turn stabilizes the training process. 
 
From above analysis, it is clear that as compared to spectral normalization, spectral regularization of Equation (\ref{key10}) encourages $W$ of the discriminator to move in a variety of directions thus preventing it from concentrating only on one direction, which in turn prevents spectral collapse. We will show in the experimental section that performing spectral regularization can indeed prevent mode collapse where spectral normalization has failed.

\section{Experiments} \label{sec:exp}

For all settings listed in Table \ref{t11}, we have conducted experiments using SN-GANs and the newly introduced spectral regularization algorithm (we use the abbreviation: SR-GANs for the spectral regularized GANs).
All procedures and settings for SN-GANs and SR-GANs are identical, except that for SR-GANs the last discriminator update implements spectral regularization (Equation \ref{key9}) and SN-GANs implement spectral normalization (Equation \ref{key4}).  

Before studying the properties of spectral regularization, we demonstrate experimentally that the new SR-GANs is superior to SN-GANs in both quality and stability.
The  Inception Score (IS) and Fr\'{e}chet Inception Distance (FID ) performances are shown in Table \ref{t22}. Please note that in the cases where mode collapse have happened, IS and FID are the best results before mode collapse. It is clearly seen that in all cases, SR-GANs outperforms SN-GANs. 
In particular, for conditional image generation on ImageNet with setting  $E_{512-64}$ and $E_{2048-64}$, SR-GAN has improved IS by 19.4\% and 44.9\%, respectively. 
On average, SR-GANs have improved the IS by 13.9\% and FID by 21.8\% over SN-GANs. 
Very importantly, in all 10 settings where mode collapse has occurred to SN-GANs,  none has happened to SR-GANs. In fact, we have not yet observed mode collapses in an extensive set of experiments.

\subsection{Conditional Generation on ImageNet}
\begin{figure*}
	\begin{subfigure}{.24\textwidth}
		\centering
		\includegraphics[height=0.7\linewidth]{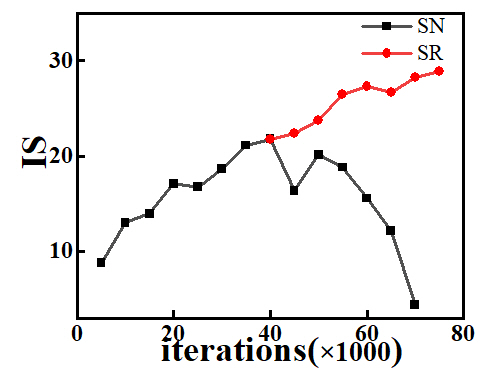}  
		\caption{\textrm{\small IS}}
	\end{subfigure}
	\begin{subfigure}{.24\textwidth}
		\centering
		\includegraphics[height=0.7\linewidth]{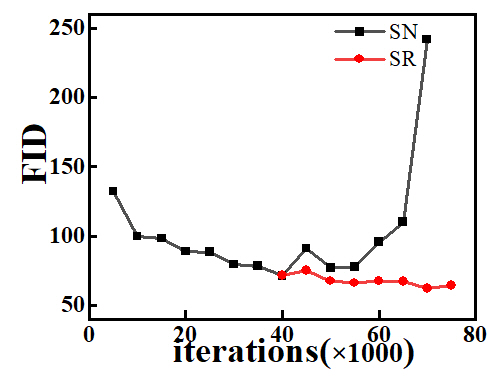}  
		\caption{\textrm{\small  FID }}
	\end{subfigure}
	\begin{subfigure}{.24\textwidth}
		\centering
		\includegraphics[height=0.7\linewidth]{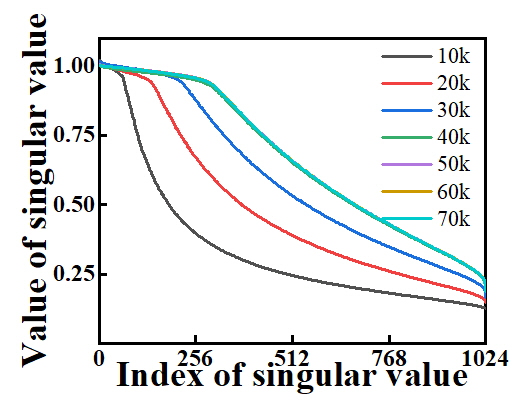}  
		\caption{\textrm{\small  Spectral Distribution }}
	\end{subfigure}
	\begin{subfigure}{.24\textwidth}
		\centering
		\includegraphics[height=0.7\linewidth]{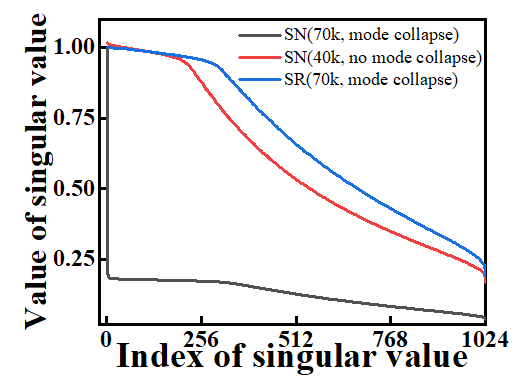}  
		\caption{\textrm{\small  Spectral Distribution }}
	\end{subfigure}
	\newline
	\begin{subfigure}{1\textwidth}
		\centering
		\includegraphics[height=0.11\linewidth]{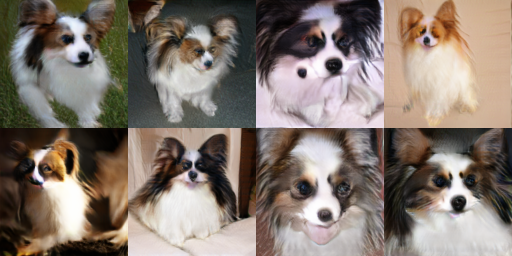}  
		\includegraphics[height=0.11\linewidth]{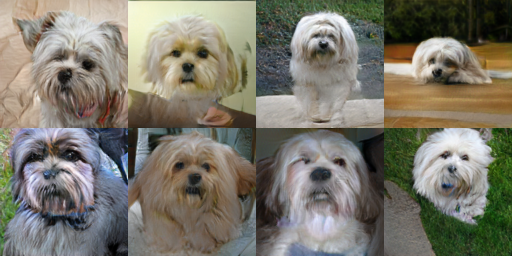} 
		\includegraphics[height=0.11\linewidth]{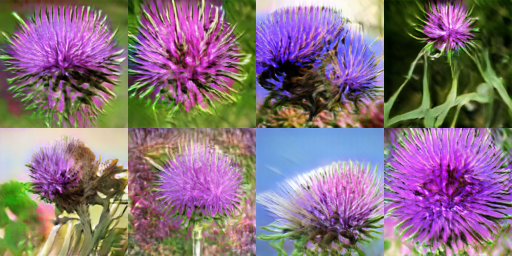}   
		\includegraphics[height=0.11\linewidth]{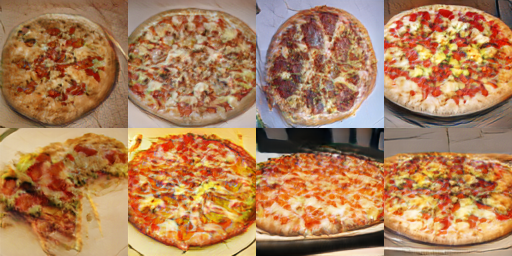} 
		\caption{\textrm{\small  SRGAN(iterations = 70k) }}
	\end{subfigure}
	\newline
	\begin{subfigure}{1\textwidth}
		\centering
		\includegraphics[height=0.11\linewidth]{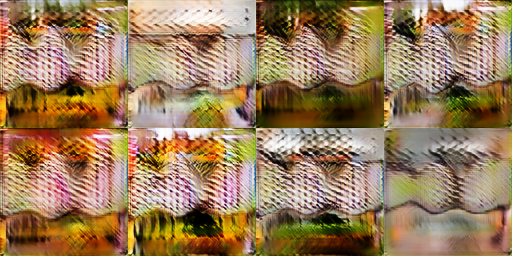}  
		\includegraphics[height=0.11\linewidth]{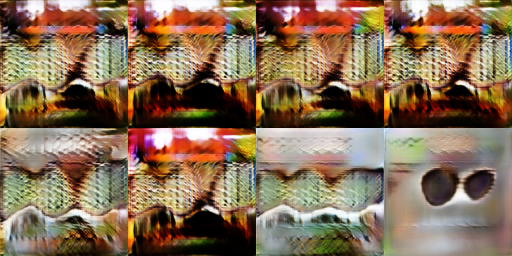} 
		\includegraphics[height=0.11\linewidth]{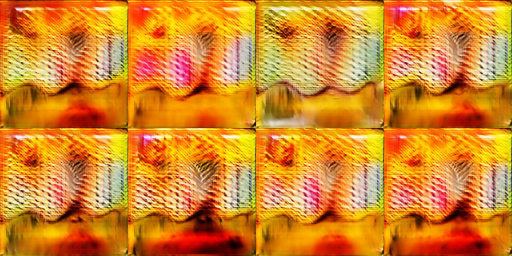}   
		\includegraphics[height=0.11\linewidth]{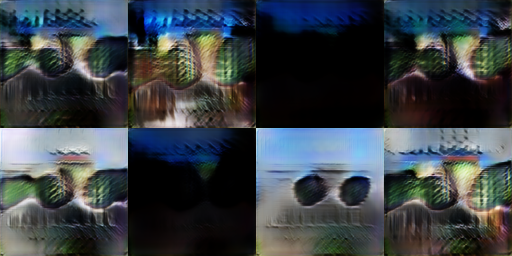} 
		\caption{\textrm{\small  SNGAN(iterations = 70k)}}
	\end{subfigure}
	\caption{Inception Score, Fr\'{e}chet Inception Distance, spectral distribution and synthetic images of SN-GAN and SR-GAN for the setting $E_{2048-64}$. (a), (b) shows the IS and FID, respectively. (c) is the spectral distributions for $layer\_16$ in SRGAN. (d) indicates the effect of SR on spectral distribution. (e), (f) are conditional generations by SRGAN and SNGAN, respectively.}
	\label{fig:fig6}
	\vspace*{-0.4cm}
	
\end{figure*}

It is clearly seen in Table \ref{t22}, spectral regularization greatly improves the performance, and contributes to stable training. For setting $E_{512-64}$, no mode collapse is observed in SN-GAN. Applying spectral regularization can further improve the performance. As for setting $E_{2048-64}$, spectral regularization avoids mode collapse, and greatly improve the image quality.

In Figure \ref{fig:fig6}, we show the training history of SN-GAN with setting $E_{2048-64}$. It is clear that at around 40k iterations, mode collapse occurs, and image quality starts to decrease dramatically. Accordingly, synthetic images in Figure \ref{fig:fig6}(f) are of low quality and limited diversity.

To demonstrate the superiority of SR over SN, we resume the training with the snapshot of SN-GAN at iterations = 40k, and apply Spectral Regularization with dynamic compensation (SR-d). The effect of SR on improving performance and guaranteeing training stability is rather obvious. Firstly, spectral regularization avoids the occurrence of mode collapse, which is supposed to happen at iterations = 40k in SN-GAN. What's more, spectral regularization contributes to performance improvement as well, i.e. IS, FID are improved by 44.9\% and 9.1\%, respectively.

Figure \ref{fig:fig6}(c) shows the spectral distribution of $layer\_16$ in SR-GAN. It is clearly shown that spectral regularization avoids spectral collapse observed in Figure \ref{fig:fig2}(j). Owing to the stronger constraint of spectral regularization on spectral distributions of each convolutional layer, SR-GANs can gain performance improvement steadily with iterations, instead of suffering from mode collapse.

\subsection{Unconditional Generation on CIFAR-10 and STL-10}

\begin{figure*}
	\begin{subfigure}{.19\textwidth}
		\centering
		\includegraphics[height=1\linewidth]{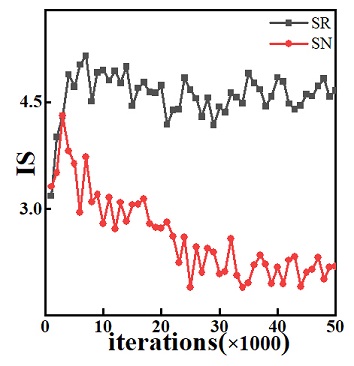}  
		\caption{\textrm{\small IS}}
	\end{subfigure}
	\begin{subfigure}{.19\textwidth}
		\centering
		\includegraphics[height=1\linewidth]{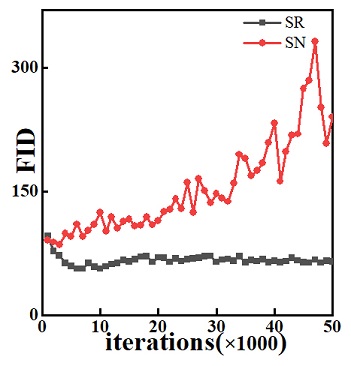}  
		\caption{\textrm{\small  FID }}
	\end{subfigure}
	\begin{subfigure}{.19\textwidth}
		\centering
		\includegraphics[height=1\linewidth]{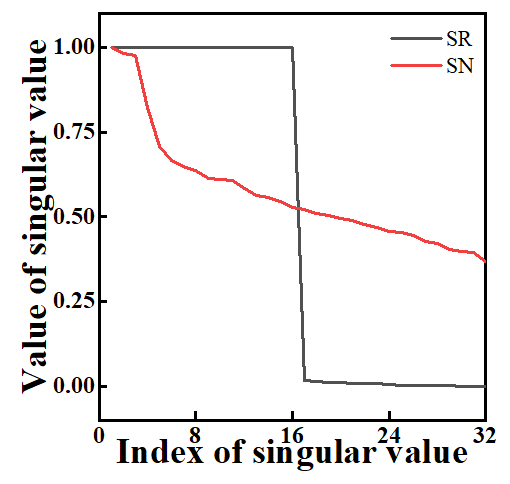}  
		\caption{\textrm{\small  Spectral Distribution }}
	\end{subfigure}
	\begin{subfigure}{.19\textwidth}
		\centering
		\includegraphics[height=1\linewidth]{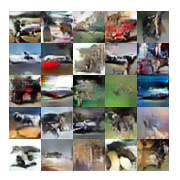}  
		\vspace*{-0.3cm}
		\caption{\textrm{\small   SR}}
	\end{subfigure}
	\begin{subfigure}{.19\textwidth}
		\centering
		\includegraphics[height=1\linewidth]{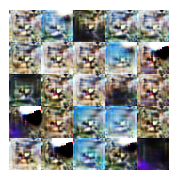}  
		\vspace*{-0.3cm}
		\caption{\textrm{\small SN}}
	\end{subfigure}
	\vspace*{-0.25cm}
	\caption{Inception Score, Fr\'{e}chet Inception Distance, spectral distribution and synthetic images of SN-GAN and SR-GAN for the setting $C_{64-32}$}
	\label{fig:fig7}
	\vspace*{-0.4cm}
	
\end{figure*}

We apply spectral Regularization with static compensation (SR-s) in unconditional generation on CIFAR-10 and STL-10 datasets. The default value of the hyperparameter $i$ for static compensation in SR-GANs is empirically set as $i = 0.5r$, where $r$ is the number of singular values in the corresponding weight matrix. Readers are referred to Appendix for the details of the network architecture settings. 

When applying spectral regularization, model performance is steadily improved, and stable training is guaranteed as shown in Table \ref{t22} .  While for spectral normalization, SN-GANs still suffers from mode collapse. 
Worst of all, when channel size is small, mode collapse will happen to SN-GAN regardless of batch size as shown in our group $C$ experiments. 
Figure \ref{fig:fig7} shows the training history of SN-GAN and SR-GAN for the setting $C_{64-32}$. It is seen that for SN-GAN, mode collapse has happened almost at the start of the training process and performance continues to deteriorate until eventually lead to mode collapse. In contrast, the performance of SR-GAN improves steadily and eventually converges (no mode collapse). 

In the comparison of spectral distribution (Figure \ref{fig:fig7}(c)), it is seen that spectral normalization cannot stop other singular values to drop significantly thus causing spectral collapse which in turn results in mode collapse. In contrast, static compensation in spectral regularization ensures that the first $i$ singular values are equal in all cases, thus ensuring that spectral collapse would not happen hence preventing mode collapse. 
Examples of generated images by the two training methods for this setting are also shown in the Figure \ref{fig:fig7}(d) and (e). It is again clearly seen that mode collapse has indeed happened to SN-GAN while the images generated by SR-GAN are of better quality and more varieties.

Through the example above, we can see that for experiment setting with too large batch size or too small channel size, SN-GANs cannot ensure stable training. Nevertheless, for various experiment setup spectral regularization can indeed guarantee training stability, and contribute to model performance, demonstrating that spectral regularization is a robust method for stable training.

\begin{figure*}
	\begin{subfigure}{.46\textwidth}
		\centering
		\includegraphics[width=1\linewidth]{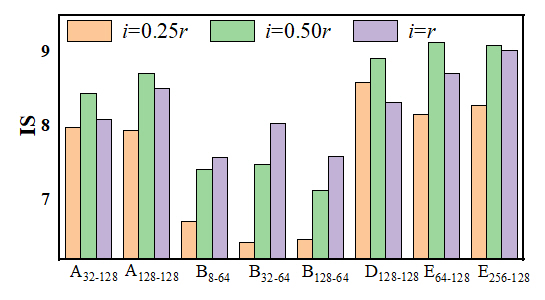}  
		\vspace*{-0.6cm}
		\caption{\textrm{\small Inception Score}}
	\end{subfigure}
	\begin{subfigure}{.46\textwidth}
		\centering
		\includegraphics[width=1\linewidth]{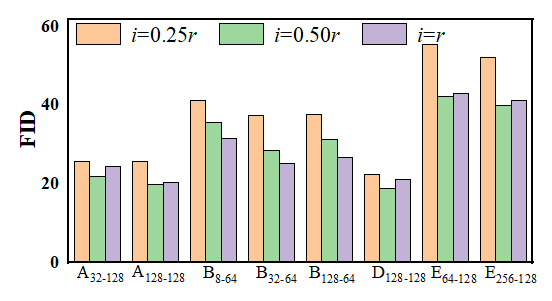}  
		\vspace*{-0.6cm}
		\caption{\textrm{\small  Fr\'{e}chet Inception Distance }}
	\end{subfigure}
	\vspace*{-0.25cm}
	\caption{The effect of $i$ on model performance. $N$ represents the number of singular values in corresponding weight matrix.}
	\label{fig:fig8}
	\vspace*{-0.0cm}
\end{figure*}	

\begin{figure}
	\begin{subfigure}{.15\textwidth}
		\centering
		\includegraphics[width=1\linewidth]{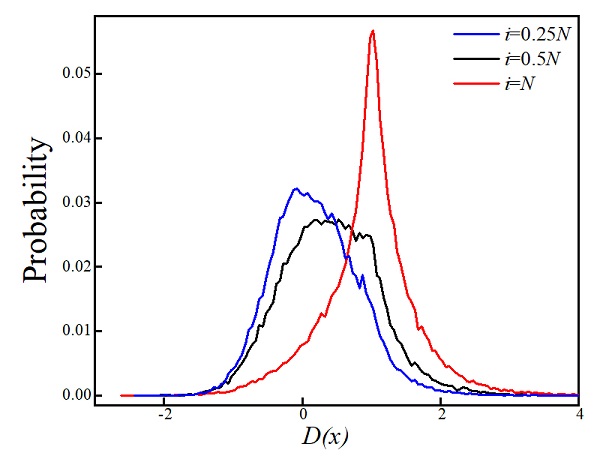}  
		\vspace*{-0.6cm}
		\caption{\textrm{\small $x \sim q_{train}$}}
	\end{subfigure}
	\begin{subfigure}{.15\textwidth}
		\centering
		\includegraphics[width=1\linewidth]{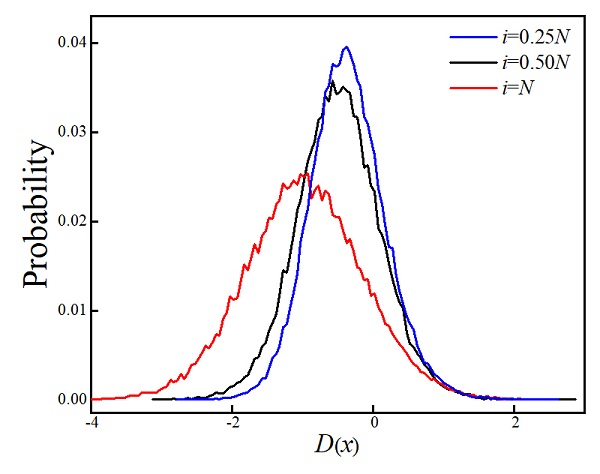}  
		\vspace*{-0.6cm}
		\caption{\textrm{\small  $x \sim q_{G}$} }
	\end{subfigure}
	\begin{subfigure}{.15\textwidth}
		\centering
		\includegraphics[width=1\linewidth]{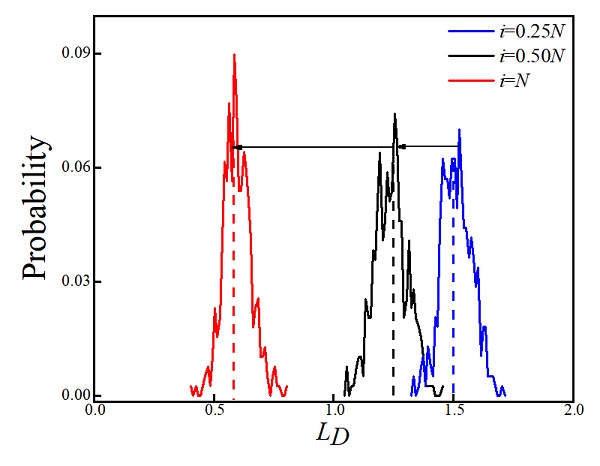}  
		\vspace*{-0.6cm}
		\caption{\textrm{\small  $L_D$}}
	\end{subfigure}
	\vspace*{-0.25cm}
	\caption{Statistics of $D(x)$ and $L_D$.}
	\label{fig:fig9}
	\vspace*{-0.0cm}
	
	\begin{subfigure}{.15\textwidth}
		\centering
		\includegraphics[width=1\linewidth]{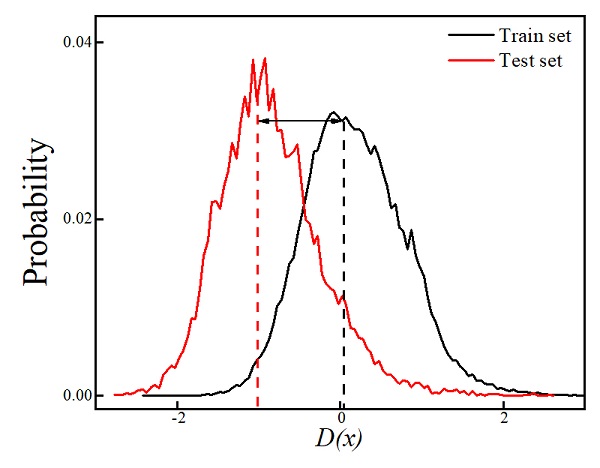}  
		\vspace*{-0.6cm}
		\caption{\textrm{\small $i$=0.25$r$}}
	\end{subfigure}
	\begin{subfigure}{.15\textwidth}
		\centering
		\includegraphics[width=1\linewidth]{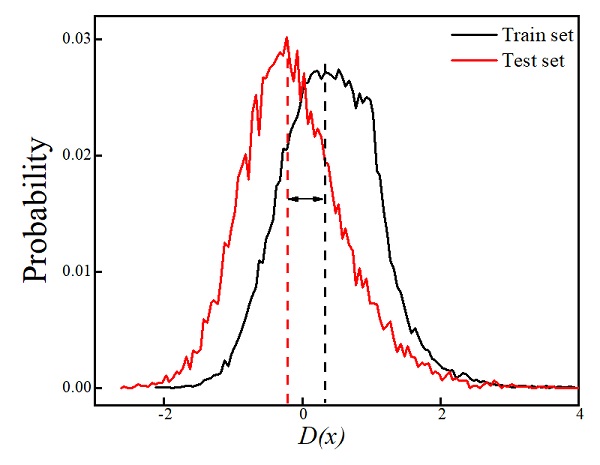}  
		\vspace*{-0.6cm}
		\caption{\textrm{\small  $i$=0.50$r$}}
	\end{subfigure}
	\begin{subfigure}{.15\textwidth}
		\centering
		\includegraphics[width=1\linewidth]{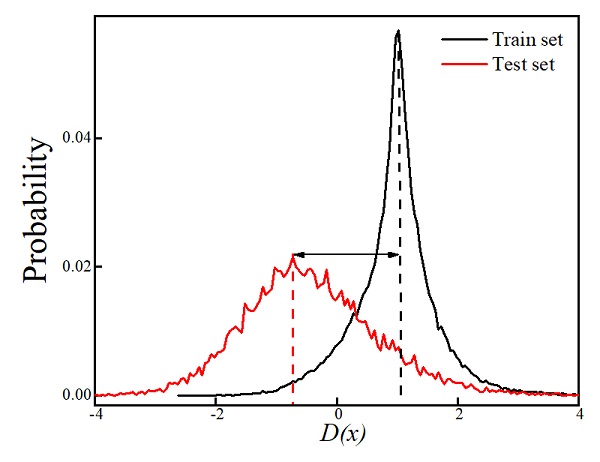}  
		\vspace*{-0.6cm}
		\caption{\textrm{\small  $i$=$r$}}
	\end{subfigure}
	\vspace*{-0.25cm}
	\caption{Statistics of $D(x)$  with setting $A_{128-128}$.}
	\label{fig:fig10}
	\vspace*{-0.0cm}
	
	\begin{subfigure}{.15\textwidth}
		\centering
		\includegraphics[width=1\linewidth]{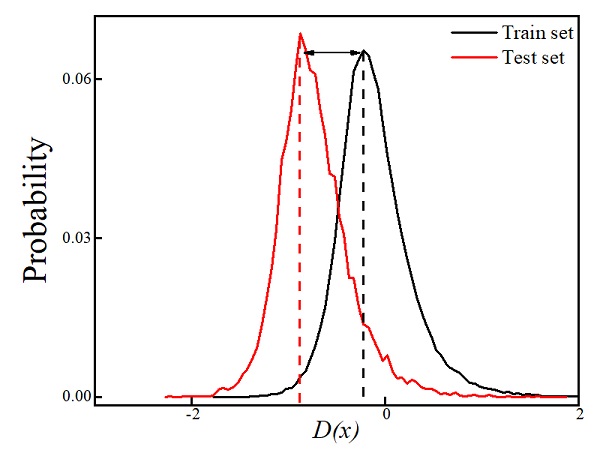}  
		\vspace*{-0.6cm}
		\caption{\textrm{\small  $i$=0.25$r$}}
	\end{subfigure}
	\begin{subfigure}{.15\textwidth}
		\centering
		\includegraphics[width=1\linewidth]{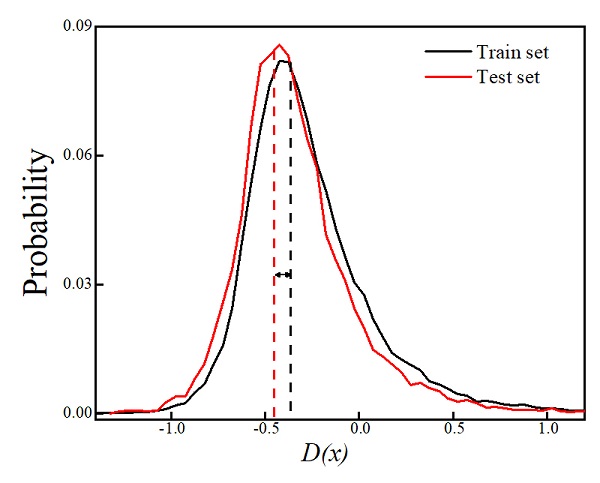}  
		\vspace*{-0.6cm}
		\caption{\textrm{\small  $i$=0.50$r$}}
	\end{subfigure}
	\begin{subfigure}{.15\textwidth}
		\centering
		\includegraphics[width=1\linewidth]{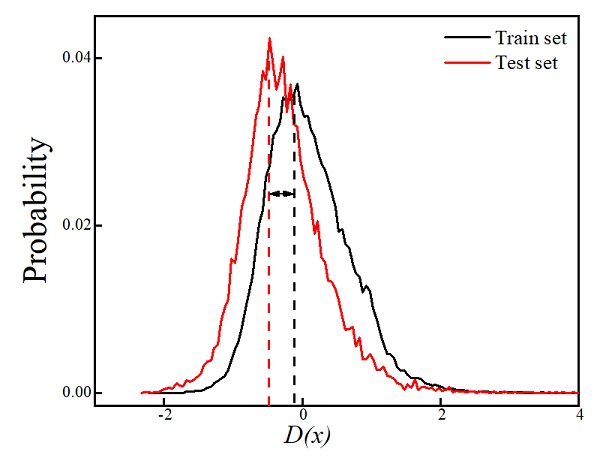}  
		\vspace*{-0.6cm}
		\caption{\textrm{\small  $i$=$r$}}
	\end{subfigure}
	\vspace*{-0.25cm}
	\caption{ Statistics of $D(x)$ with setting $B_{128-64}$.}
	\label{fig:fig11}
	\vspace*{-0.2cm}
\end{figure}

\subsection{The Hyperparameter $i$ in SR-GANs} \label{sec4.2}

When applying static compensation in spectral regularization, there is a single hyperparameter $i$, and its value will affect performances. In the experiments above, $i$ in SR-GANs is set to $i=0.5r$, where $r$ is the number of singular values. Clearly, when $i=1$, SR-GAN is the same as SN-GAN, therefore SN-GAN is a special case of SR-GAN. To investigate the effect of $i$, we gradually increase $i$, and observe its influence on model performance. 
In Figure \ref{fig:fig8}, we show the Inception Scores and Fr\'{e}chet Inception Distances for different values of $i$. For experiment groups $A, D$ and $E$, increasing $i$ from 0.25$N$ to 0.5$N$, the performances are improved. However, continuously increasing $i$ from 0.5$N$ to $N$, the performances deteriorate. For experiments in group $B$, performances increase steadily with $i$.

To understand why $i$ affects performances in this way,  we feed the discriminator function with the generated data and real data from both the training and testing sets, and then record the statistics of $D(x)$ in Equation (\ref{key2}) and the discriminator objective $L_D$ in Equation (\ref{key5}). For explanation convenience, some typical results are illustrated here. 

The probability distributions of $D(x)$ for the generated data $D(x)|_{x \sim q_{G}}$ and that for the training data  $D(x)|_{x \sim q_{train}}$ for the setting of $A_{128-128}$ and different $i$ values are shown in Figure \ref{fig:fig9} (a) and  Figure \ref{fig:fig9} (b), respectively. Here  $q_{train}$ represents training set, and  $q_{G}$ represents generated set. The probability distributions of $L_D$ is shown in Figure \ref{fig:fig9} (c). 

When increasing $i$ from 0.25$N$ to $N$, the distributions of $D(x)|_{x \sim q_{train}}$ have a tendency of moving to the right, and at the same time the distributions of $D(x)|_{x \sim q_{G}}$ have a tendency of moving to the left. This means that the discriminator can better discriminate between the real and generated samples. This is also verified by the distributions of $L_D$ as can be clearly seen in Figure \ref{fig:fig9} (c). 

To investigate discriminator's performance on the testing set, we show the probability distributions of  $D(x)|_{x \sim q_{train}}$ and $D(x)|_{x \sim q_{test}}$ for the setting $A_{128-128}$ in Figure \ref{fig:fig10}, where $q_{test}$ represents test set. It is seen that for $i=0.25r$ and $i=0.5r$, the two distributions are more similar to each other than that of $i=r$. In the case of $i=r$, the discriminator behaves significantly differently between the training data and testing data, this means that overfitting has occurred and results in a drop in performances.  In summary, Figure \ref{fig:fig9} and Figure \ref{fig:fig10} explain the performance drop for setting  $i=r$ in experiment groups $A, D$ and $E$.

Furthermore, we monitor the statistics of $D(x)$ for the settings in group $B$  to explain why $i$ affects the behaviors of SR-GANs as in Figure \ref{fig:fig8}.  The probability distributions of $D(x)$  for the setting $B_{128-64}$ are shown in Figure \ref{fig:fig11}. We can see that for all the $i$ values, the probability distributions of the discriminator output for the training and testing data agree well with each other, indicating no overfitting has occurred. 

Although there is no systematic method for determining the best $i$ value for different settings, our experiences is that setting $i=0.5r$ seems to work well. In a series of extensive experiments we conducted,  setting $i=0.5r$, SR-GANs always outperform SN-GANs and very importantly, we have not yet observed mode collapse. 

\begin{figure*}
	\begin{subfigure}{.46\textwidth}
		\centering
		\includegraphics[width=1\linewidth]{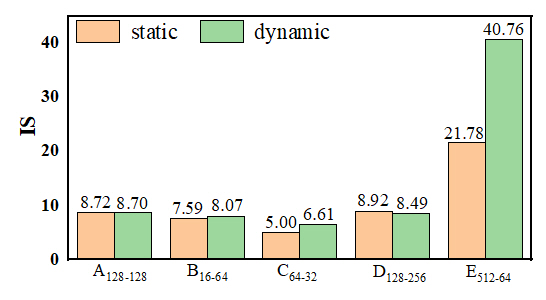}  
		\vspace*{-0.6cm}
		\caption{\textrm{\small Inception Score}}
	\end{subfigure}
	\begin{subfigure}{.46\textwidth}
		\centering
		\includegraphics[width=1\linewidth]{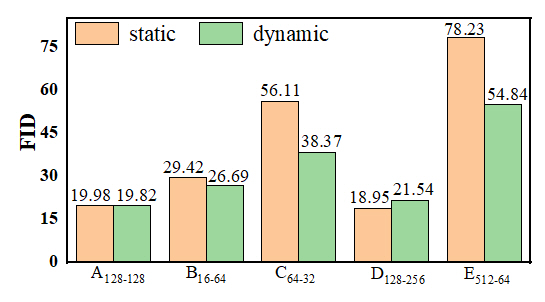}  
		\vspace*{-0.6cm}
		\caption{\textrm{\small  Fr\'{e}chet Inception Distance }}
	\end{subfigure}
	\vspace*{-0.25cm}
	\caption{Comparison of two compensation methods.}
	\label{fig:fig12}
	\vspace*{-0.4cm}
\end{figure*}

\subsection{Static Compensation vs Dynamic Compensation}
Two kinds of compensation method are proposed in this paper: static and dynamic compensation. In Figure \ref{fig:fig6}(d) and Figure \ref{fig:fig7}(c), we show how these two compensation methods affect the spectral distribution.

Static compensation encourages to use the first $i$ singular values. Therefore, spectral distributions mainly concentrate on the first $i$ singular values, leaving the rest almost zero (Figure \ref{fig:fig7}(c)). Meanwhile, the shape of spectral distributions are totally different from that in SN-GANs. Unlike static compensations, which try to carry out compensation at the largest scale, dynamic compensations focus on guaranteeing the shapes of spectral distributions, and avoiding the occurrence of spectral collapse. As we can see in Figure \ref{fig:fig6}(c), the shapes of spectral distributions hardly change at the end of training.

Recalling the Equation \ref{key66}, static compensations averagely utilize the first $i$ singular vectors, regardless of their differences. However, dynamic compensations consider the differences, and carry out compensation adaptively.
To further show the difference between static and dynamic compensation, comparative trials are conducted based on settings $A_{128-128}$, $B_{16-64}$, $C_{64-32}$, $D_{128-256}$ and $E_{512-64}$. For these setting, two kinds of compensation methods are applied, and results are shown in Figure \ref{fig:fig12}. 
static compensations achieve better results on unconditional generation with setting $A_{128-128}$
and $D_{128-256}$. While dynamic compensations behave better in the rest tasks. Therefore, no convincing evidence shows dynamic compensation is better than static compensation, or the other way around. Our suggestion is that, for image generation on datasets of low resolutions and few categories, static compensation may works better. For image generation on datasets of high resolutions and more categories, dynamic compensation is better.

%
%
%

\section{Conclusions}

In this paper, we monitor spectral distributions of the discriminator's  weight matrices in SN-GANs. We discover that when mode collapse occurs to a SN-GAN, a large number of its weight matrices singular values will drop to very small values, and we introduce the concept of spectral collapse to describe this phenomenon. We have provided strong evidence to link mode collapse with spectral collapse. Based on such link, we have successfully developed a spectral regularization technique for training GANs. We show that by compensating the spectral distributions of the  weight matrices, we can successfully prevent spectral collapse which in turn can successfully prevent mode collapse. In a series of extensive experiments, we have successfully demonstrated that preventing spectral collapse can not only avoid mode collapse but also can improve GANs performances.

\appendices
\section{Proof of Corollary 1}
\noindent \textbf{Corollary 1.} If a linear function $f = Wx$ satisfies Lipschitz constraint: $\left \| f(x_1) - f(x_2) \right \| \leq \left \| x_1 - x_2 \right \| $, then the supremum of the Lipschitz constraint is obtained when all the singular values of the weight matrix $W$ are 1.

\noindent Proof: 
Because $f$ is a linear function: $f(x)= Wx$. The 1-Lipschitz constraint for $f$ can be expressed as: 
\begin{equation} \label{keya1}
\left \|Wx  \right \| \leqslant \left \|x  \right \|
\end{equation}
Equation \ref{keya1}  is equivalent to:
\begin{equation}
\left \|Wx  \right \|^{2} \leqslant \left \|x  \right \|^{2}
\end{equation} 
and,
\begin{equation}
\left \|Wx  \right \|^{2}= x^TW^TWx=x^TV \Sigma  V^Tx
\end{equation}
where columns of $V$, $[v_1, \cdots, v_n]$ are eigenvectors of $W^TW$, and diagonal entries of diagonal matrix $\Sigma $ are eigenvalues of $W^TW$. 

Taking $y=V^Tx$, then 
\begin{equation}
\left \|Wx  \right \|^{2}=y^T\Sigma  y =\lambda_1y_1^2 +\cdots +\lambda_ny_n^2
\end{equation}
where $\lambda_{i}$ is the $i$-th eigenvalue, and $y_i$ is the $i$-th element of $y$.

Because $W^TW$ is symmetric, $V^T=V^{-1}$, then
\begin{equation}
\left \|y  \right \|
^2=y^Ty=x^TVV^Tx=x^Tx=\left \|x  \right \|^2
\end{equation}
Finally, $\left \|Wx  \right \|^{2} \leqslant \left \|x  \right \|^{2}$ is equivalent to  $\lambda_1y_1^2 +\cdots +\lambda_ny_n^2 \leqslant y_1^2 +\cdots +y_n^2$. We can see that the upper bound of 1-Lipschitz constraint can be obtained only when all eigenvalues of $W^TW$ are 1. In other words, all the singular values of $W$ are 1. 

\section{Architecture and Optimization Settings}
In this paper, we employ SN-GAN architecture for image generation task. To better illustrate how we change the channel size in the discriminator architecture, we show the architecture details in Figure \ref{fig:figa1}. 
The weight in the convolutional layer is in the format [$out$, $in$, $h$, $w$], where $out$ is the output channel,  $in$ represents the input channel, $h$ and $w$ are kernel sizes.

Particularly, there are 10 convolutional layers ($layer\_0 \sim layer\_9$) in $D$ network for image generation on CIFAR-10 and STL-10, and 17 convolutional layers  ($layer\_0 \sim layer\_16$) in $D$ network for image generation on ImageNet.
ch in Figure \ref{fig:figa1} corresponds to channel size of discriminator function in main text, where extensive experiments are conducted with different settings of ch. All the experiments are conducted based on the following architecture. Image generation on STL-10 shares the same architecture with that on CIFAR-10. Thus, images in STL-10 are compressed to 32 $\times$ 32 pixels, identical to the resolution of images in CIFAR-10. For conditional generation on ImageNet, images are compressed to 128 $\times$ 128 pixels.

The optimization settings follow SN-GANs. To be specific, for image generation on CIFAR-10 and STL-10, the learning rate is taken as 0.0002, the number of updates of the discriminator per one update of the generator $n_{critic}$ is 5, and Adam optimizer is used as the optimization with the first and second order momentum parameters as 0 and 0.9, respectively. For image generation on ImageNet, the learning rate for $G$ and $D$ is taken as 0.0001 and 0.0004, respectively, the number of updates of the discriminator per one update of the generator $n_{critic}$ is 1. Adam optimizer is used as the optimization with the first and second order momentum parameters as 0 and 0.9, respectively. And, spectral normalization is applied in $G$ architecture. To alleviate the huge demand for computational facility, we apply the trick of gradient accumulation, which is proposed in the implementation of BigGANs.

\section{Spectral Distribution}

\begin{figure*}
	\begin{subfigure}[b]{.46\textwidth}
		\centering
		\includegraphics[width=0.8\linewidth]{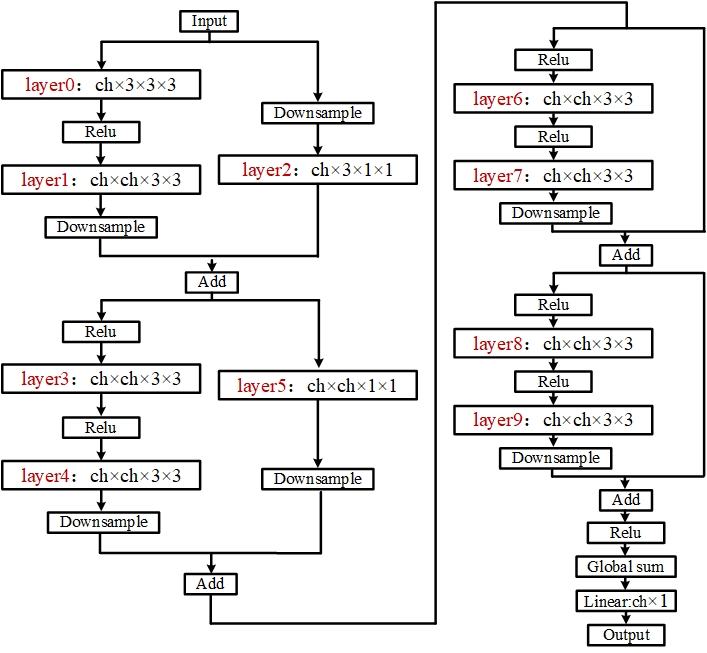}  
		\caption{\textrm{\small $D$ architecture for CIFAR-10 and STL-10}}
	\end{subfigure}
	\begin{subfigure}[b]{.46\textwidth}
		\centering
		\includegraphics[width=0.8\linewidth]{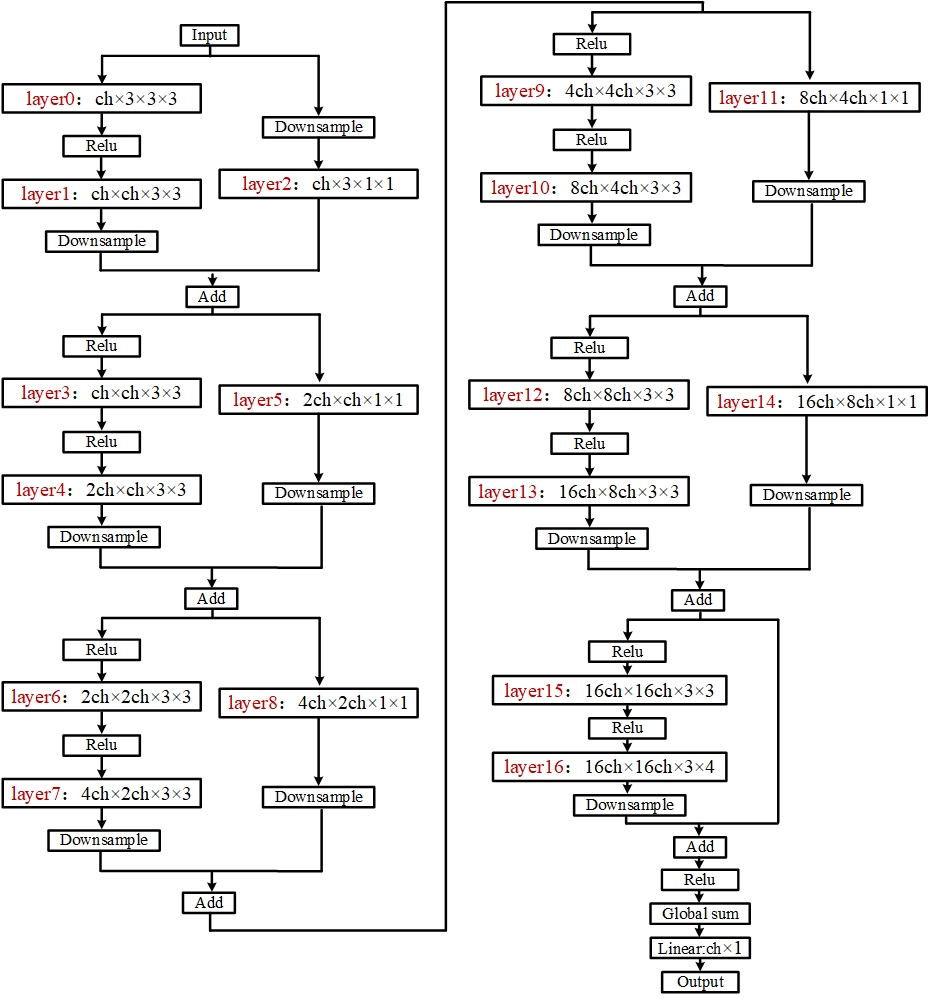}  
		\caption{\textrm{\small  $D$ architecture for ImageNet}}
	\end{subfigure}
	\vspace*{-0.25cm}
	\caption{Architecture of $D$.}
	\label{fig:figa1}
\end{figure*}

In Figure \ref{fig:figa3} $\sim$ Figure \ref{fig:figa10}, we show the spectral distributions for settings, which suffer from mode collapse.  
We can see that spectral collapse and mode collapse always go side by side. 
For image generation on CIFAR-10 and STL-10, spectral collapse is observed in all layers except  $layer\_2$ and $layer\_5$, which act as the role of skip connection, as shown in Figure \ref{fig:figa1}. While for image generation on ImageNet, spectral collapse is observed  in all layers except $lay\_2$.

%
%

%
%



%

\bibliographystyle{IEEEtran}
\bibliography{IEEEexample}

%






\begin{figure*}[h]
	\begin{subfigure}{.12\textwidth}
		\centering
		\includegraphics[width=0.95\linewidth]{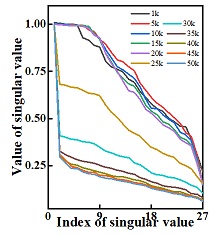}  
		\caption{$layer\_0$}
	\end{subfigure}
	\begin{subfigure}{.12\textwidth}
		\centering
		\includegraphics[width=0.95\linewidth]{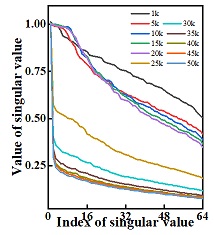}  
		\caption{$layer\_1$}
	\end{subfigure}
	\begin{subfigure}{.12\textwidth}
		\centering
		\includegraphics[width=0.95\linewidth]{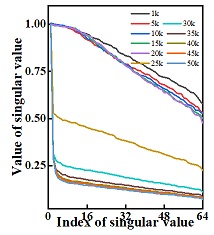}  
		\caption{$layer\_3$}
	\end{subfigure}
	\begin{subfigure}{.12\textwidth}
		\centering
		\includegraphics[width=0.95\linewidth]{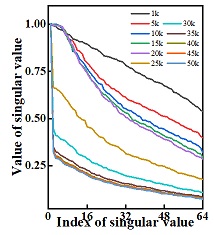}  
		\caption{$layer\_4$}
	\end{subfigure}
	\begin{subfigure}{.12\textwidth}
		\centering
		\includegraphics[width=0.95\linewidth]{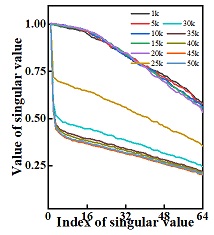}  
		\caption{$layer\_6$}
	\end{subfigure}
	\begin{subfigure}{.12\textwidth}
		\centering
		\includegraphics[width=0.95\linewidth]{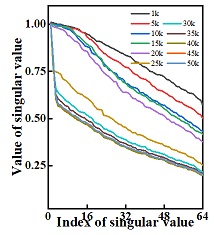}  
		\caption{$layer\_7$}
	\end{subfigure}
	\begin{subfigure}{.12\textwidth}
		\centering
		\includegraphics[width=0.95\linewidth]{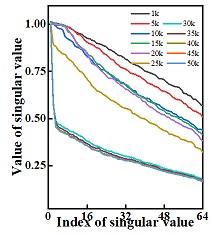}  
		\caption{$layer\_8$}
	\end{subfigure}
	\begin{subfigure}{.12\textwidth}
		\centering
		\includegraphics[width=0.95\linewidth]{a2f.jpg}  
		\caption{$layer\_9$}
	\end{subfigure}
	\vspace*{-0.25cm}
	\caption{Spectral distributions  settings  $B_{64-64}$.}
	\label{fig:figa2}
	\vspace*{-0.0cm}
\end{figure*}	

\begin{figure*}[h]
	\begin{subfigure}{.12\textwidth}
		\centering
		\includegraphics[width=0.95\linewidth]{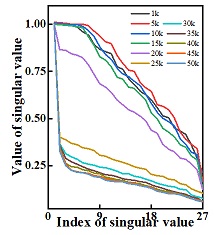}  
		\caption{$layer\_0$}
	\end{subfigure}
	\begin{subfigure}{.12\textwidth}
		\centering
		\includegraphics[width=0.95\linewidth]{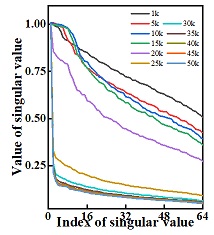}  
		\caption{$layer\_1$}
	\end{subfigure}
	\begin{subfigure}{.12\textwidth}
		\centering
		\includegraphics[width=0.95\linewidth]{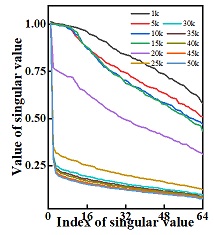}  
		\caption{$layer\_3$}
	\end{subfigure}
	\begin{subfigure}{.12\textwidth}
		\centering
		\includegraphics[width=0.95\linewidth]{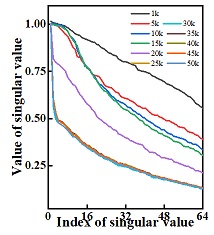}  
		\caption{$layer\_4$}
	\end{subfigure}
	\begin{subfigure}{.12\textwidth}
		\centering
		\includegraphics[width=0.95\linewidth]{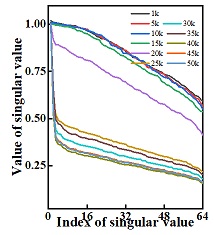}  
		\caption{$layer\_6$}
	\end{subfigure}
	\begin{subfigure}{.12\textwidth}
		\centering
		\includegraphics[width=0.95\linewidth]{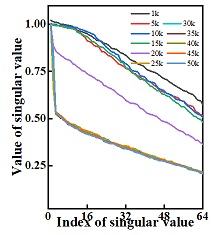}  
		\caption{$layer\_7$}
	\end{subfigure}
	\begin{subfigure}{.12\textwidth}
		\centering
		\includegraphics[width=0.95\linewidth]{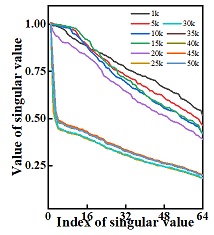}  
		\caption{$layer\_8$}
	\end{subfigure}
	\begin{subfigure}{.12\textwidth}
		\centering
		\includegraphics[width=0.95\linewidth]{a3f.jpg}  
		\caption{$layer\_9$}
	\end{subfigure}
	\vspace*{-0.25cm}
	\caption{Spectral distributions  settings  $B_{128-64}$.}
	\label{fig:figa3}
	\vspace*{-0.0cm}
\end{figure*}	

\begin{figure*}[h]
	\begin{subfigure}{.12\textwidth}
		\centering
		\includegraphics[width=0.95\linewidth]{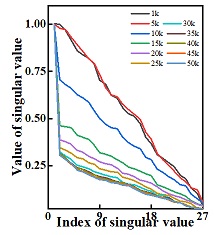}  
		\caption{$layer\_0$}
	\end{subfigure}
	\begin{subfigure}{.12\textwidth}
		\centering
		\includegraphics[width=0.95\linewidth]{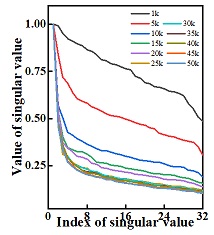}  
		\caption{$layer\_1$}
	\end{subfigure}
	\begin{subfigure}{.12\textwidth}
		\centering
		\includegraphics[width=0.95\linewidth]{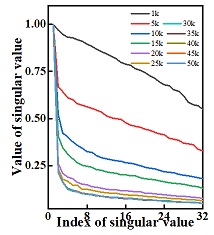}  
		\caption{$layer\_3$}
	\end{subfigure}
	\begin{subfigure}{.12\textwidth}
		\centering
		\includegraphics[width=0.95\linewidth]{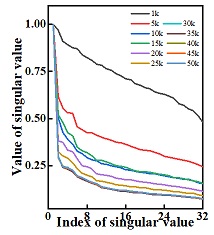}  
		\caption{$layer\_4$}
	\end{subfigure}
	\begin{subfigure}{.12\textwidth}
		\centering
		\includegraphics[width=0.95\linewidth]{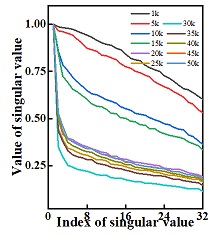}  
		\caption{$layer\_6$}
	\end{subfigure}
	\begin{subfigure}{.12\textwidth}
		\centering
		\includegraphics[width=0.95\linewidth]{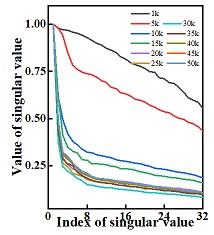}  
		\caption{$layer\_7$}
	\end{subfigure}
	\begin{subfigure}{.12\textwidth}
		\centering
		\includegraphics[width=0.95\linewidth]{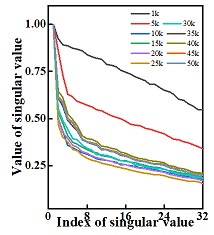}  
		\caption{$layer\_8$}
	\end{subfigure}
	\begin{subfigure}{.12\textwidth}
		\centering
		\includegraphics[width=0.95\linewidth]{a4f.jpg}  
		\caption{$layer\_9$}
	\end{subfigure}
	\vspace*{-0.25cm}
	\caption{Spectral distributions  settings  $C_{8-32}$.}
	\label{fig:figa4}
	\vspace*{-0.0cm}
\end{figure*}

\begin{figure*}[h]
	\begin{subfigure}{.12\textwidth}
		\centering
		\includegraphics[width=0.95\linewidth]{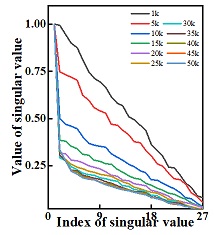}  
		\caption{$layer\_0$}
	\end{subfigure}
	\begin{subfigure}{.12\textwidth}
		\centering
		\includegraphics[width=0.95\linewidth]{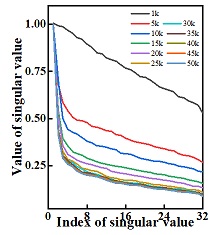}  
		\caption{$layer\_1$}
	\end{subfigure}
	\begin{subfigure}{.12\textwidth}
		\centering
		\includegraphics[width=0.95\linewidth]{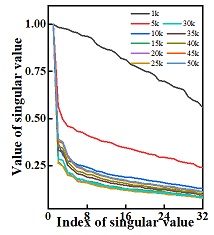}  
		\caption{$layer\_3$}
	\end{subfigure}
	\begin{subfigure}{.12\textwidth}
		\centering
		\includegraphics[width=0.95\linewidth]{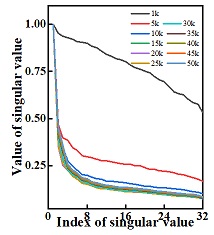}  
		\caption{$layer\_4$}
	\end{subfigure}
	\begin{subfigure}{.12\textwidth}
		\centering
		\includegraphics[width=0.95\linewidth]{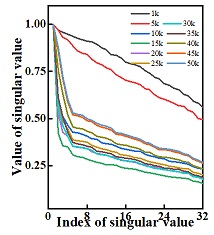}  
		\caption{$layer\_6$}
	\end{subfigure}
	\begin{subfigure}{.12\textwidth}
		\centering
		\includegraphics[width=0.95\linewidth]{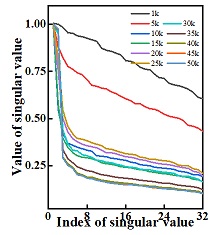}  
		\caption{$layer\_7$}
	\end{subfigure}
	\begin{subfigure}{.12\textwidth}
		\centering
		\includegraphics[width=0.95\linewidth]{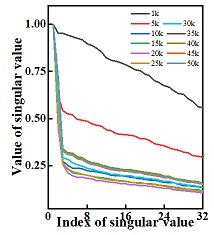}  
		\caption{$layer\_8$}
	\end{subfigure}
	\begin{subfigure}{.12\textwidth}
		\centering
		\includegraphics[width=0.95\linewidth]{a5f.jpg}  
		\caption{$layer\_9$}
	\end{subfigure}
	\vspace*{-0.25cm}
	\caption{Spectral distributions  settings  $C_{16-32}$.}
	\label{fig:figa5}
	\vspace*{-0.0cm}
\end{figure*}

\begin{figure*}[h]
	\begin{subfigure}{.12\textwidth}
		\centering
		\includegraphics[width=0.95\linewidth]{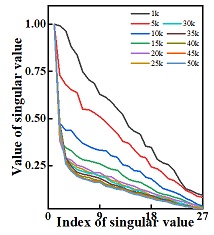}  
		\caption{$layer\_0$}
	\end{subfigure}
	\begin{subfigure}{.12\textwidth}
		\centering
		\includegraphics[width=0.95\linewidth]{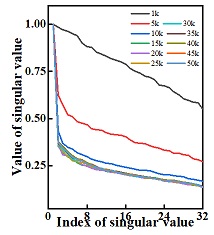}  
		\caption{$layer\_1$}
	\end{subfigure}
	\begin{subfigure}{.12\textwidth}
		\centering
		\includegraphics[width=0.95\linewidth]{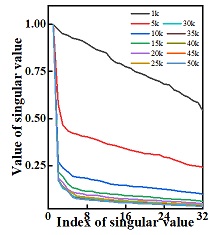}  
		\caption{$layer\_3$}
	\end{subfigure}
	\begin{subfigure}{.12\textwidth}
		\centering
		\includegraphics[width=0.95\linewidth]{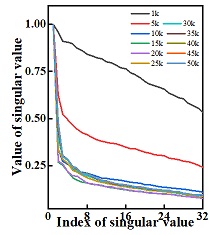}  
		\caption{$layer\_4$}
	\end{subfigure}
	\begin{subfigure}{.12\textwidth}
		\centering
		\includegraphics[width=0.95\linewidth]{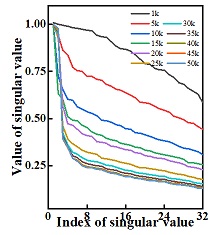}  
		\caption{$layer\_6$}
	\end{subfigure}
	\begin{subfigure}{.12\textwidth}
		\centering
		\includegraphics[width=0.95\linewidth]{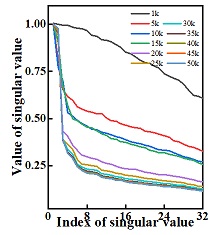}  
		\caption{$layer\_7$}
	\end{subfigure}
	\begin{subfigure}{.12\textwidth}
		\centering
		\includegraphics[width=0.95\linewidth]{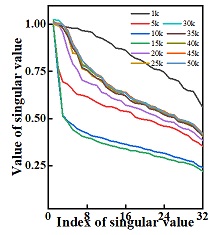}  
		\caption{$layer\_8$}
	\end{subfigure}
	\begin{subfigure}{.12\textwidth}
		\centering
		\includegraphics[width=0.95\linewidth]{a6f.jpg}  
		\caption{$layer\_9$}
	\end{subfigure}
	\vspace*{-0.25cm}
	\caption{Spectral distributions  settings  $C_{32-32}$.}
	\label{fig:figa6}
	\vspace*{-0.0cm}
\end{figure*}

\begin{figure*}[h]
	\begin{subfigure}{.12\textwidth}
		\centering
		\includegraphics[width=0.95\linewidth]{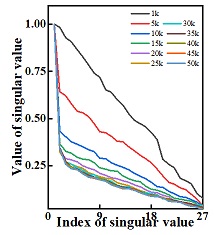}  
		\caption{$layer\_0$}
	\end{subfigure}
	\begin{subfigure}{.12\textwidth}
		\centering
		\includegraphics[width=0.95\linewidth]{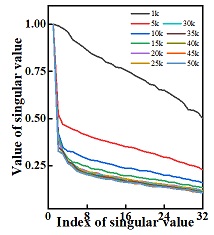}  
		\caption{$layer\_1$}
	\end{subfigure}
	\begin{subfigure}{.12\textwidth}
		\centering
		\includegraphics[width=0.95\linewidth]{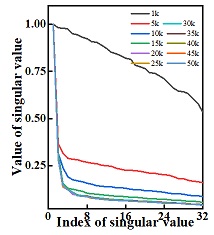}  
		\caption{$layer\_3$}
	\end{subfigure}
	\begin{subfigure}{.12\textwidth}
		\centering
		\includegraphics[width=0.95\linewidth]{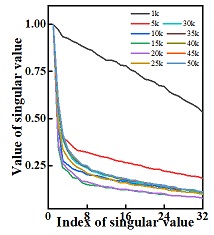}  
		\caption{$layer\_4$}
	\end{subfigure}
	\begin{subfigure}{.12\textwidth}
		\centering
		\includegraphics[width=0.95\linewidth]{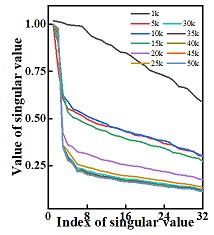}  
		\caption{$layer\_6$}
	\end{subfigure}
	\begin{subfigure}{.12\textwidth}
		\centering
		\includegraphics[width=0.95\linewidth]{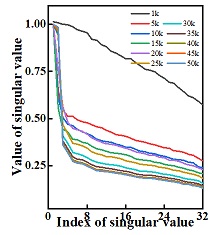}  
		\caption{$layer\_7$}
	\end{subfigure}
	\begin{subfigure}{.12\textwidth}
		\centering
		\includegraphics[width=0.95\linewidth]{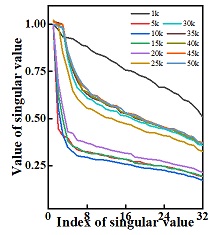}  
		\caption{$layer\_8$}
	\end{subfigure}
	\begin{subfigure}{.12\textwidth}
		\centering
		\includegraphics[width=0.95\linewidth]{a7f.jpg}  
		\caption{$layer\_9$}
	\end{subfigure}
	\vspace*{-0.25cm}
	\caption{Spectral distributions  settings  $C_{64-32}$.}
	\label{fig:figa7}
	\vspace*{-0.0cm}
\end{figure*}

\begin{figure*}[h]
	\begin{subfigure}{.12\textwidth}
		\centering
		\includegraphics[width=0.95\linewidth]{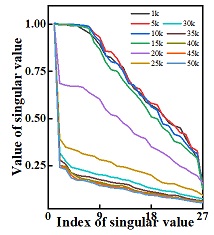}  
		\caption{$layer\_0$}
	\end{subfigure}
	\begin{subfigure}{.12\textwidth}
		\centering
		\includegraphics[width=0.95\linewidth]{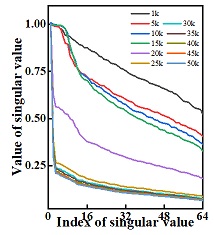}  
		\caption{$layer\_1$}
	\end{subfigure}
	\begin{subfigure}{.12\textwidth}
		\centering
		\includegraphics[width=0.95\linewidth]{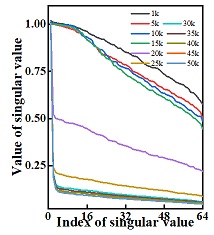}  
		\caption{$layer\_3$}
	\end{subfigure}
	\begin{subfigure}{.12\textwidth}
		\centering
		\includegraphics[width=0.95\linewidth]{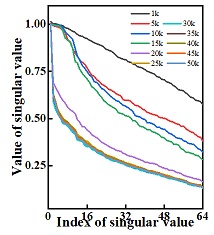}  
		\caption{$layer\_4$}
	\end{subfigure}
	\begin{subfigure}{.12\textwidth}
		\centering
		\includegraphics[width=0.95\linewidth]{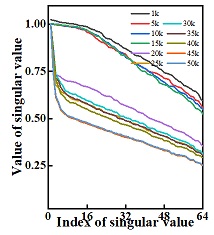}  
		\caption{$layer\_6$}
	\end{subfigure}
	\begin{subfigure}{.12\textwidth}
		\centering
		\includegraphics[width=0.95\linewidth]{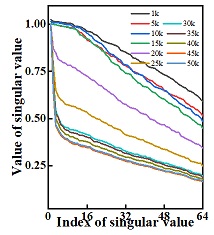}  
		\caption{$layer\_7$}
	\end{subfigure}
	\begin{subfigure}{.12\textwidth}
		\centering
		\includegraphics[width=0.95\linewidth]{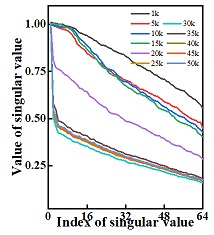}  
		\caption{$layer\_8$}
	\end{subfigure}
	\begin{subfigure}{.12\textwidth}
		\centering
		\includegraphics[width=0.95\linewidth]{a8f.jpg}  
		\caption{$layer\_9$}
	\end{subfigure}
	\vspace*{-0.25cm}
	\caption{Spectral distributions  settings  $E_{256-64}$.}
	\label{fig:figa8}
	\vspace*{-0.0cm}
\end{figure*}

\begin{figure*}[h]
	\begin{subfigure}{.12\textwidth}
		\centering
		\includegraphics[width=0.95\linewidth]{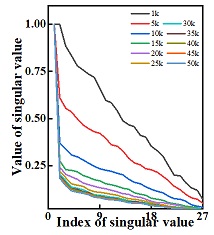}  
		\caption{$layer\_0$}
	\end{subfigure}
	\begin{subfigure}{.12\textwidth}
		\centering
		\includegraphics[width=0.95\linewidth]{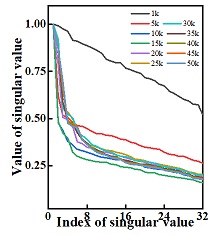}  
		\caption{$layer\_1$}
	\end{subfigure}
	\begin{subfigure}{.12\textwidth}
		\centering
		\includegraphics[width=0.95\linewidth]{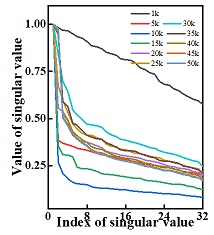}  
		\caption{$layer\_3$}
	\end{subfigure}
	\begin{subfigure}{.12\textwidth}
		\centering
		\includegraphics[width=0.95\linewidth]{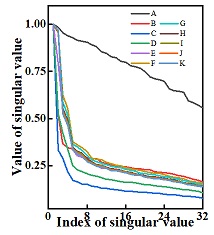}  
		\caption{$layer\_4$}
	\end{subfigure}
	\begin{subfigure}{.12\textwidth}
		\centering
		\includegraphics[width=0.95\linewidth]{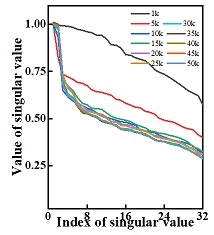}  
		\caption{$layer\_6$}
	\end{subfigure}
	\begin{subfigure}{.12\textwidth}
		\centering
		\includegraphics[width=0.95\linewidth]{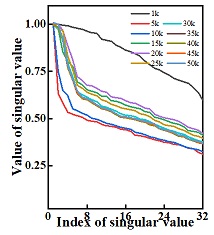}  
		\caption{$layer\_7$}
	\end{subfigure}
	\begin{subfigure}{.12\textwidth}
		\centering
		\includegraphics[width=0.95\linewidth]{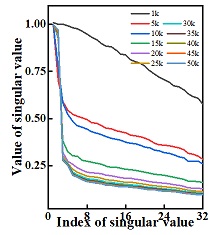}  
		\caption{$layer\_8$}
	\end{subfigure}
	\begin{subfigure}{.12\textwidth}
		\centering
		\includegraphics[width=0.95\linewidth]{a9f.jpg}  
		\caption{$layer\_9$}
	\end{subfigure}
	\vspace*{-0.25cm}
	\caption{Spectral distributions  settings  $E_{256-32}$.}
	\label{fig:figa9}
	\vspace*{-0.0cm}
\end{figure*}

\begin{figure*}[h]
	\begin{subfigure}{.12\textwidth}
		\centering
		\includegraphics[width=0.95\linewidth]{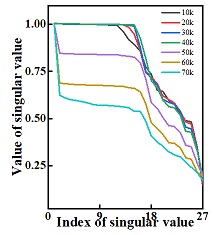}  
		\caption{$layer\_0$}
	\end{subfigure}
	\begin{subfigure}{.12\textwidth}
		\centering
		\includegraphics[width=0.95\linewidth]{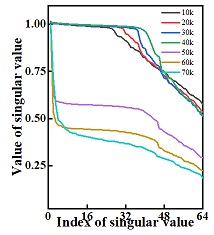}  
		\caption{$layer\_1$}
	\end{subfigure}
	\begin{subfigure}{.12\textwidth}
		\centering
		\includegraphics[width=0.95\linewidth]{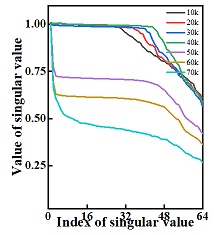}  
		\caption{$layer\_3$}
	\end{subfigure}
	\begin{subfigure}{.12\textwidth}
		\centering
		\includegraphics[width=0.95\linewidth]{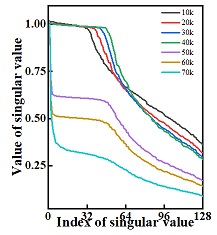}  
		\caption{$layer\_4$}
	\end{subfigure}
	\begin{subfigure}{.12\textwidth}
		\centering
		\includegraphics[width=0.95\linewidth]{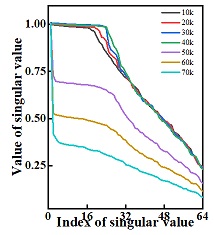}  
		\caption{$layer\_6$}
	\end{subfigure}
	\begin{subfigure}{.12\textwidth}
		\centering
		\includegraphics[width=0.95\linewidth]{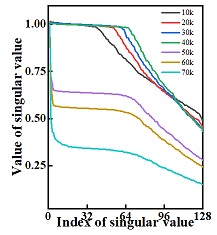}  
		\caption{$layer\_7$}
	\end{subfigure}
	\begin{subfigure}{.12\textwidth}
		\centering
		\includegraphics[width=0.95\linewidth]{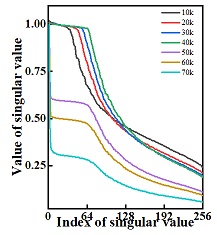}  
		\caption{$layer\_8$}
	\end{subfigure}
	\begin{subfigure}{.12\textwidth}
		\centering
		\includegraphics[width=0.95\linewidth]{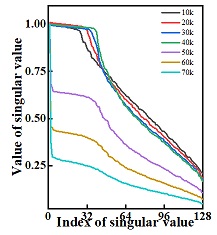}  
		\caption{$layer\_9$}
	\end{subfigure}
	\newline
	\begin{subfigure}{.12\textwidth}
		\centering
		\includegraphics[width=0.95\linewidth]{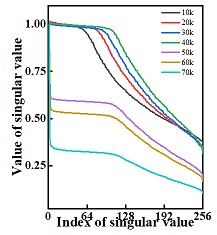}  
		\caption{$layer\_0$}
	\end{subfigure}
	\begin{subfigure}{.12\textwidth}
		\centering
		\includegraphics[width=0.95\linewidth]{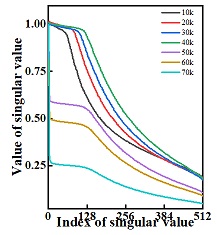}  
		\caption{$layer\_1$}
	\end{subfigure}
	\begin{subfigure}{.12\textwidth}
		\centering
		\includegraphics[width=0.95\linewidth]{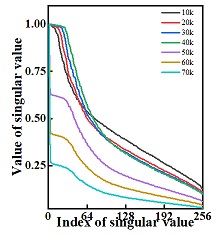}  
		\caption{$layer\_3$}
	\end{subfigure}
	\begin{subfigure}{.12\textwidth}
		\centering
		\includegraphics[width=0.95\linewidth]{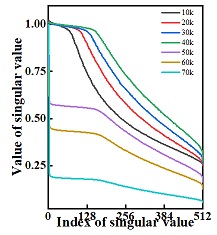}  
		\caption{$layer\_4$}
	\end{subfigure}
	\begin{subfigure}{.12\textwidth}
		\centering
		\includegraphics[width=0.95\linewidth]{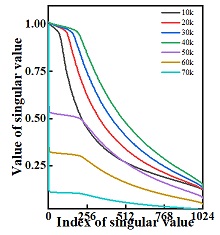}  
		\caption{$layer\_6$}
	\end{subfigure}
	\begin{subfigure}{.12\textwidth}
		\centering
		\includegraphics[width=0.95\linewidth]{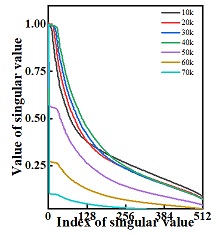}  
		\caption{$layer\_7$}
	\end{subfigure}
	\begin{subfigure}{.12\textwidth}
		\centering
		\includegraphics[width=0.95\linewidth]{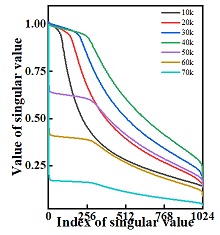}  
		\caption{$layer\_8$}
	\end{subfigure}
	\begin{subfigure}{.12\textwidth}
		\centering
		\includegraphics[width=0.95\linewidth]{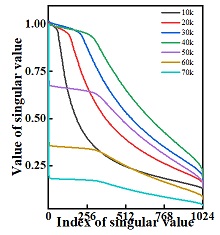}  
		\caption{$layer\_9$}
	\end{subfigure}
	\vspace*{-0.25cm}
	\caption{Spectral distributions  settings  $E_{2048-64}$.}
	\label{fig:figa10}
	\vspace*{-0.0cm}
\end{figure*}

\end{document}